\documentclass[lettersize,journal]{IEEEtran}
\IEEEoverridecommandlockouts

\usepackage{dutchcal}
\usepackage{amsmath}
\usepackage{bbding}
\usepackage{amsmath,amssymb,amsfonts}
\usepackage{algorithmicx}
\usepackage{algorithm}
\usepackage{graphicx}
\usepackage{textcomp}
\usepackage{xcolor}
\usepackage{booktabs}
\usepackage{comment}
\usepackage{multirow}
\usepackage{multicol}
\newcommand{\tabincell}[2]{\begin{tabular}{@{}#1@{}}#2\end{tabular}}
\usepackage{algpseudocode}
\usepackage{breakurl}
\usepackage{epstopdf}
\usepackage{pbox}
\usepackage{siunitx}
\usepackage{enumerate}
\usepackage[hidelinks]{hyperref}
\usepackage{alltt}
\usepackage{makecell}
\usepackage{pifont}
\usepackage{graphicx}
\usepackage{cite}
\usepackage{picinpar}
\usepackage{amsmath}
\usepackage{url}
\usepackage{flushend}
\usepackage{colortbl}
\usepackage{soul}
\usepackage{multirow}
\usepackage[caption=false,font=normalsize,labelfont=sf,textfont=sf]{subfig}
\usepackage{booktabs}
\usepackage{scalerel}
\usepackage{tikz}
\usetikzlibrary{svg.path}

\definecolor{orcidlogocol}{HTML}{A6CE39}
\tikzset{
	orcidlogo/.pic={
		\fill[orcidlogocol] svg{M256,128c0,70.7-57.3,128-128,128C57.3,256,0,198.7,0,128C0,57.3,57.3,0,128,0C198.7,0,256,57.3,256,128z};
		\fill[white] svg{M86.3,186.2H70.9V79.1h15.4v48.4V186.2z}
		svg{M108.9,79.1h41.6c39.6,0,57,28.3,57,53.6c0,27.5-21.5,53.6-56.8,53.6h-41.8V79.1z M124.3,172.4h24.5c34.9,0,42.9-26.5,42.9-39.7c0-21.5-13.7-39.7-43.7-39.7h-23.7V172.4z}
		svg{M88.7,56.8c0,5.5-4.5,10.1-10.1,10.1c-5.6,0-10.1-4.6-10.1-10.1c0-5.6,4.5-10.1,10.1-10.1C84.2,46.7,88.7,51.3,88.7,56.8z};
	}
}

\newcommand\orcidicon[1]{\href{https://orcid.org/#1}{\mbox{\scalerel*{
				\begin{tikzpicture}[yscale=-1,transform shape]
				\pic{orcidlogo};
				\end{tikzpicture}
			}{|}}}}

\usepackage{hyperref} %<--- Load after everything else

\usepackage{scalerel}

\begin{document}

\title{PiDAn: A Coherence Optimization Approach for Backdoor Attack Detection and Mitigation in Deep Neural Networks}

\author{Yue Wang$^{1}$ \orcidicon{0000-0002-9260-3970}, \IEEEmembership{Graduate Student Member, IEEE}, Wenqing Li$^{2}$ \orcidicon{0000-0001-5612-9899}, \IEEEmembership{Member, IEEE}, Esha Sarkar$^{1}$ \orcidicon{0000-0002-4473-7368}, Muhammad Shafique$^{3}$ \orcidicon{??}, \IEEEmembership{Senior Member, IEEE}, \IEEEmembership{Graduate Student Member, IEEE}, Michail Maniatakos$^{4}$ \orcidicon{0000-0001-6899-0651}, \IEEEmembership{Senior Member, IEEE}, and Saif Eddin Jabari$^{5}$ \orcidicon{0000-0002-2314-5312}, \IEEEmembership{Senior Member, IEEE}% <-this % stops a space
	\thanks{$^{1}$Yue Wang and Esha Sarkar are with the Department of Electrical and Computer Engineering, Tandon school of Engineering, New York University, Brooklyn, NY, USA {\tt\small yw3576@nyu.edu}, {\tt\small esha.sarkar@nyu.edu}}%
	\thanks{$^{2}$Wenqing Li is with the Division of Engineering, New York University Abu Dhabi, Abu Dhabi, UAE 
		{\tt\small wl54@nyu.edu}}%
    \thanks{$^{3}$Muhammad Shafique is with the Division of Engineering, New York University Abu Dhabi, Abu Dhabi, UAE and the Department of Electrical and Computer Engineering, Tandon School of Engineering, New York University, Brooklyn, NY, USA
		{\tt\small ms12713@nyu.edu}}%
	\thanks{$^{4}$Corresponding author. Michail Maniatakos is with the Division of Engineering, New York University Abu Dhabi, Abu Dhabi, UAE and the Department of Electrical and Computer Engineering, Tandon School of Engineering, New York University, Brooklyn, NY, USA
		{\tt\small mihalis.maniatakos@nyu.edu}}%
	\thanks{$^{5}$Saif Eddin Jabari is with the Division of Engineering, New York University Abu Dhabi, Abu Dhabi, UAE and the Department of Civil and Urban Engineering, Tandon School of Engineering, New York University, Brooklyn, NY, USA
		{\tt\small sej7@nyu.edu}}%
}

\markboth{IEEE Transactions on Neural Networks and Learning Systems,~Vol.~14, No.~8, Mar~2022}%
{Shell \MakeLowercase{\textit{et al.}}: PiDAn: A Coherence Optimization Approach for Backdoor Attack Detection and Mitigation in Deep Neural Networks}

\maketitle

\begin{abstract}
 
% Backdoor attacks impose new threat in deep learning models, where a backdoor or trojan is inserted into the deep neural network by poisoning the training dataset and the images carrying the adversary trigger will be assigned the target label. The challenge of defending these attacks lies on that only the attacker knows the secret trigger and the target, especially when the trigger representations fuse with the authentic ones. It lacks a comprehensive backdoor detection and mitigation method when the trigger representations show less distinction with the authentic ones.
 
% To this end, we analyze how representations are affected by backdoor attacks and propose a coherence optimization based poison purifying (CoPP) algorithm. Our tentative research shows that representations of poisoned data and authentic data in target class are still embedded in different linear subspaces, which implies they show different coherence with some latent spaces. Based on this, the proposed CoPP algorithm learns a sample-wise weight vector to maximize the projected coherence of weighted samples, where we demonstrate that the learned weight vector has a natural `grouping effect' and is distinguishable between authentic data and poisoned data, which enables the detection and mitigation of backdoor attacks. Based on the theoretical analysis and extensive experimental results, we demonstrate the effectiveness of CoPP in defending against backdoor attacks that use
% different settings of poison samples on GTSRB and ILSVRC2012 datasets.  

%NEW abstract Mihalis
Backdoor attacks impose a new threat in Deep Neural Networks (DNNs), where a backdoor is inserted into the neural network by poisoning the training dataset, misclassifying inputs that contain the adversary trigger. The major challenge for defending against these attacks is that only the attacker knows the secret trigger and the target class. The problem is further exacerbated by the recent introduction of ``Hidden Triggers,'' where the triggers are carefully fused into the input, bypassing detection by human inspection and causing backdoor identification through anomaly detection to fail.

To defend against such imperceptible attacks, in this work we systematically analyze how representations, i.e., the set of neuron activations for a given DNN when using the training data as inputs, are affected by backdoor attacks. We propose \textbf{PiDAn}, an algorithm based on coherence optimization purifying the poisoned data. Our analysis shows that representations of poisoned data and authentic data in the target class are still embedded in different linear subspaces, which implies that they show different coherence with some latent spaces. Based on this observation, the proposed PiDAn algorithm learns a sample-wise weight vector to maximize the projected coherence of weighted samples, where we demonstrate that the learned weight vector has a natural "grouping effect" and is distinguishable between authentic data and poisoned data. This enables the systematic detection and mitigation of backdoor attacks. Based on our theoretical analysis and experimental results, we demonstrate the effectiveness of PiDAn in defending against backdoor attacks that use different settings of poisoned samples on GTSRB and ILSVRC2012 datasets. Our PiDAn algorithm can detect more than 90\% infected classes and identify 95\% poisoned samples.

%To defend against such imperceptible attacks, we systematically analyze how representations are affected by backdoor attacks, and propose a \textbf{C}oherence \textbf{O}ptimization-based \textbf{P}oison \textbf{Pur}ifying (COPPur) algorithm. Our analysis shows that representations of poisoned data and authentic data in the target class are still embedded in different linear subspaces, which implies they show different coherence with some latent spaces. Based on this observation, the proposed COPPur algorithm learns a sample-wise weight vector to maximize the projected coherence of weighted samples, where we demonstrate that the learned weight vector has a natural ``grouping effect'' and is distinguishable between authentic data and poisoned data, which enables the detection and mitigation of backdoor attacks. Based on theoretical analysis and experimental results, we demonstrate the effectiveness of COPPur in defending against backdoor attacks that use different settings of poison samples on GTSRB and ILSVRC2012 datasets. Our COPPur algorithm can detect more than 90\% infected classes and identify 95\% poisoned samples. 

\end{abstract}

\begin{IEEEkeywords}
Deep Neural Networks (DNNs), Backdoor Attacks, Backdoor Defense, Optimization, Machine Learning Security 
\end{IEEEkeywords}

\section{Introduction and Related Work}
\IEEEPARstart{B}{ackdoors} in Deep Neural Networks (DNNs) are a recent insidious attack \cite{dumford2020backdooring, zhong2020backdoor, Rakin_2020_CVPR, li2020invisible, yao2019latent, bagdasaryan2020blind}. These attacks aim to insert a backdoor inside of the target DNN model so that the attacker can control the model output (typically the class label) when triggered using a specific input pattern. {Data poisoning is an effective way to implement this attack, where the poisoned samples with the attacker-chosen properties (e.g., patched with specific triggers) are added to the genuine training dataset and labeled as the target class. After training on the mixture of genuine and poisoned data, the backdoored model can connect the attacker-chosen properties to the target label and also hold high accuracy on clean test samples.}
Since its discovery in 2017, there has been a plethora of research with attack literature focusing on developing stealthier attacks and defense literature developing schemes to protect models against those specific attacks. The defense mechanisms typically analyze the properties of backdoor triggers (attack signatures) to distinguish between benign and malicious instances. In this work, we take a different approach in comparison to the state-of-the-art defenses, shifting the focus to fundamental properties of DNNs instead of trigger-specific properties.
%The defenses focus on detecting the state of the art stealthy attacks, their algorithms leverage the properties (or attack signatures) of the existing ones. 
%Therefore, a new attack, with different attack signatures may evade the defenses. 
%Intuitively, a defense mechanism would be robust and generalizable if it leverages the fundamental properties of neural networks to understand the difference between malicious and benign instances. 

The major challenge of developing a generalizable defense is that the trigger space is infinite, making its analysis cumbersome. 
In DNNs, deeper layers can extract underlying features of complex data, which transforms the input data associated to different objects to linearly separable representations, i.e., the set of neuron activations for a given DNN when using the training data as inputs. The features associated to different objects lie in different latent linear subspaces \cite{cohen2020separability, brahma2015deep}. This is a fundamental property of DNNs. In a backdoored network, even though the poisoned data is manually assigned the label of one object, their features are different, i.e., intuitively the representations for poisoned data and the genuine/clean data lie in two different linear subspaces. Defense mechanisms leveraging the inherent DNN characteristics have the added advantage of being attack-agnostic, i.e., their detection capability can be consistent across different datasets as well as different attack algorithms. 

\begin{table*}[]
    \centering
    \caption{Comparison between our defense method and state-of-the-art defense methods. SCAn, Spectral Signature, Activation Clustering and our proposed method defend against backdoors by only analyzing the training data.}
    \vspace{-0.1in}
    %\footnotesize
    \begin{tabular}{c|c|c|c|c|c||c|c|c|c}
    \hline
    \hline
    & \tabincell{c}{Fuzz \& \\Majority \\~\cite{sarkar2020backdoor}} & \tabincell{c}{Neural \\cleanse \\~\cite{wang2019neural}} & \tabincell{c}{ABS~\cite{liu2019abs}} & \tabincell{c}{Strip~\cite{gao2019strip}} & \tabincell{c}{SentiNet \\~\cite{chou2020sentinet}} & \tabincell{c}{SCAn \\~\cite{tang2021demon}} & \tabincell{c}{Spectral\\ signature \\~\cite{tran2018spectral}} & \tabincell{c}{Activation\\ clustering \\~\cite{chen2018detecting}} & \tabincell{c}{\bf PiDAn \\ \bf (This work)} \\
    \hline
    Partial backdoor defense & &  &  &  &  & \checkmark  & \checkmark  & \checkmark & \checkmark \\
    \hline
    Hidden trigger defense & &  &  &  &  & \checkmark  &   &  & \checkmark \\
    \hline
    Backdoor detection & & \checkmark & \checkmark & \checkmark & \checkmark & \checkmark  &  & \checkmark & \checkmark \\
    \hline
    Model recovery & \checkmark & \checkmark &  &  &  &  &  \checkmark & \checkmark & \checkmark \\
    \hline
    \hline
    \end{tabular}
    \vspace{-0.2in}
    \label{related work}
\end{table*}

Initial work on backdooring neural networks studied the feasibility of backdoor attacks using small patches as triggers \cite{gu2019badnets, black_badnets, liu2020reflection}. Following that line of work, researchers identified that these attacks are based on a strong connection between the trigger and the output, and formed mitigation strategies for detecting the backdoors based on the trigger dominance. In order to break the dominance of triggers, Tang et al., \cite{tang2021demon} proposed to not only add patched data to the target classes, but also add covering data (data from non-source classes patched with triggers) to their original classes. In this manner, only the output of the trigger-carrying source data can be flipped by the triggers. They also show that, in their case, the representations of the authentic data and the poisoned data overlap significantly, which makes it difficult to detect the poisoned data. The more advanced clean-label backdoor attack~\cite{shafahi2018poison} and hidden trigger backdoor attack~\cite{saha2020hidden} generate poisoned samples that can be naturally labeled as the target class, i.e., they look like samples from the target class. %In this manner, the triggers are not revealed in the training process.

The existence of backdoors in a model is not reflected in the regular performance metrics like test accuracy. 
%The challenge of detecting backdoor attacks lies in the fact that only the attacker knows the trigger and the attack target. At the same time, the backdoored model maintains its intended functionality in the absence of the trigger, thus it is hard to identify backdoors by monitoring the test accuracy. 
Therefore, defense literature focuses on analyzing the potential trigger properties (e.g., strong connection with the output labels) in order to perform backdoor detection \cite{kolouri2020universal, qiao2019defending, achille2018emergence, huang2020one, xiang2020detection, xiang2019benchmark, guo2019tabor}. Defenses can be broadly classified in two categories: (1) Methods that focus on the connection between the trigger and the output, and (2) Methods that focus on distinguishing poisoned samples from clean samples in the training set. We first summarize the functionality of the state-of-the-art defense methods in Table~\ref{related work}, and then further analyze each category. 

With regards to the defense literature that examines the strong connection between the trigger and the output, the backdoor can be detected by restoring a small pattern that can cause any image to be misclassified as the target \cite{wang2019neural}, or the small region on image that is significantly important for classification \cite{huang2019neuroninspect}, or small range of neurons stimulated by triggers \cite{liu2019abs}.  More specifically, Neural Cleanse \cite{wang2019neural} and Deepinspect \cite{chen2019deepinspect} detect the existence of backdoor by reverse-engineering the trigger pattern that can misclassify all images attached with this trigger pattern to each target class. For the backdoored model, the trigger pattern for the true infected class will be much smaller than that of the uninfected ones in magnitude. This type of defense can discover the trigger by accessing the victim models. However, it is computationally expensive to reverse engineer the trigger pattern for each class, especially for a large number of classes. Furthermore, their performance is also constrained by the size of the triggers.
ABS \cite{liu2019abs} assumes that a set of inner neurons within the certain range will be triggered since the trigger representations will dominate the misclassification. They detect the poisonous neurons by stimulating the clean images. However, this method shows less effectiveness for some emerging attacks \cite{saha2020hidden, tang2021demon} which break the dominance of the trigger. STRIP \cite{gao2019strip} and SentiNet \cite{chou2020sentinet} aim to detect poisonous images in the testing phase. These two methods hold the same limitation as ABS, which relies on the dominance of triggers for misclassification. Finally, authors in \cite{sarkar2020backdoor} fuzz the input and perform majority voting to suppress the backdoor. This method, however, is not capable of detecting the existence of backdoors in the model, and incurs performance overhead for clean and backdoored models indistinguishably. 
%\vspace{-0.05in}
\begin{figure*}[htbp!]
	\centering
    \subfloat[]{\includegraphics[scale=0.5]{{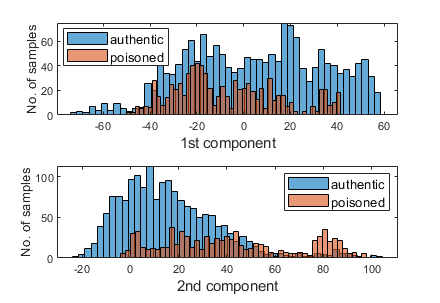}}\label{intr1}}
    \hfil
    \subfloat[]{\includegraphics[scale=0.5]{{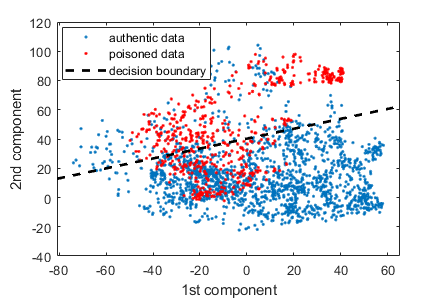}}\label{intr2}}
    %\vspace{-0.2in}
    \caption{Illustration of not well shaped features/components: (a) distributions of first component and second component and (b) classification results by SVM.}
    \vspace{-0.2in}
    \label{intr}
\end{figure*}

The above defense methods are effective when all images with triggers are misclassified. Therefore, in order to detect "partial" backdoors, where only images from some classes are triggered, recent literature focuses on the fact that the poisoned data and authentic data receive the same output class for different reasons \cite{chen2018detecting}, thus, their learned representations show different characteristics. These methods \cite{chen2018detecting,tang2021demon,tran2018spectral} utilize anomaly detection techniques \cite{li2017linearity,feng2019slow,qin2021iiot,yu2021moninet} to detect the existence of poisoned samples with the assumption that they depart from their non-poisoned counterparts, especially in the representation space~\cite{chen2018detecting}, and they are still effective even if the trigger is fused with other clean features for  classification. 
%After correctly identifying the poisoned data, it is easy to mitigate the influence of backdoor. %Current methods in this group rely on the significant separation between the representations of poisoned and clean data.
Within this category, Activation Clustering (AC) \cite{chen2018detecting} and SCAn \cite{tang2021demon} detect the infected class by distinguishing the poisoned and authentic samples. AC focuses on the Independent Component Analysis (ICA) principal components of the representations for each class. They assume that the top principal components of the data in the infected class fall into two separable clusters, the poisoned one and the authentic one. Then they can detect the infected class and access the poisoned samples for repairing the malicious model.  %However, this defense can be bypassed when the representations of the poisonous data and authentic data are designed to be less distinguishable in top features, as shown in \cite{tang2019demon, saha2020hidden}. 
%However, this defense may lose efficiency in detecting the infected class for complicated attack tasks as shown in \cite{tang2021demon, saha2020hidden}.
Instead of directly distinguishing two clusters of representations, 
%SCAn provides a strategy to distinguish the distributions of the poisonous and the authentic.
SCAn recovers the representation distribution for each class. The representations from the authentic class follow single Gaussian distribution and representations in the infected class follow two Gaussian mixture (the authentic and the poisoned).
It should be noted that, to simplify the problem, they assume the poisoned and the authentic representations in the infected classes both follow the Gaussian distributions with the same variance and different means. Thus, the difference between the poisoned and authentic distributions relies on their different means. 
%However, they did not evaluate the ability of their method for repairing the infected models. 
Spectral signature~\cite{tran2018spectral} also falls into this category. It removes the samples having large correlations with the top principal components of the corrupted data, which in turn have high probability of departing distribution of the authentic data. Despite of the effectiveness of above methods, they might face challenges when learned features are not well shaped, i.e., they don't have clear clusters. To illustrate this, we present the distributions of the top principal components with respect to poisoned and authentic representations in Figure~\ref{intr1}, where the representations are obtained by feeding the poisoned dataset (from GTSRB, under hidden trigger attack) into the DNN. In Figure~\ref{intr2}, we also present the classification results via support vector machine (SVM) based on the principal components as the core technique of above methods is classification/clustering upon learned features. For SVM, we assume the labels of poisoned and authentic representations are known. It is noted that undesired classification results are obtained, e.g., too many misclassified points, since the principal components are not well separated. This implies traditional defense methods might fail in these scenarios.
%\vspace{-0.1in}

To this end, we design a novel poisoned data analysis based method, \textbf{PiDAn}, which utilizes data coherence to extract more abstract and discriminant features, i.e., a weight vector, based on which authentic and poisoned representations can be clearly indicated. Poison purifying (e.g., detection and mitigation of backdoor attacks) is further realized by implementing likelihood ratio test based algorithm and clustering technique. Our novel contributions are summarized as follows: 
\begin{enumerate}
    \item We study how backdoor attacks affect representations of DNNs, where our findings show that the representations of authentic data and  poisoned data in the target class are embedded in different linear subspaces, which implies they showcase different coherence with the latent spaces.
    \item We propose the PiDAn algorithm, which is based on a coherence optimization problem. It learns a sample-wise weight vector via maximizing the projected coherence of weighted samples. The learned weight vector is demonstrated to have a natural "grouping effect" and is distinguishable between authentic data and poisoned data, which implies that the PiDAn algorithm can successfully detect and mitigate backdoors.
    \item We test our method on state-of-the-art neural networks, e.g., Alexnet, with wide range of visual recognition applications, from general object identification to specific traffic sign recognition, using basic and advanced trigger designs. We also compare our method with state-of-the-art methods, i.e., Spectral Signature, Activation Clustering and SCAn, to demonstrate the efficacy and benefits of our proposed method in detecting infected classes and identifying poisoned samples.
\end{enumerate}

\section{Threat Model}
\label{threat}
Our threat model 
%, similar to \cite{saha2020hidden, gu2019badnets, tang2021demon}, 
considers an attacker who can manipulate the training dataset by adding poisoned samples, but does not interfere with the training process and/or the model parameters. Such an attack scenario has been adopted by recent work such as Hidden Trigger \cite{saha2020hidden}, TaCT \cite{tang2021demon}, and Badnets \cite{gu2019badnets}. 
%This threat is introduced with the requirement of plenty of data for building effective deep learning models. 
In this scenario, attacks can originate from the data source provider (i.e. data coming from untrusted sources), or from intruders or insiders who can stealthily inject poisoned samples into the collected training set \cite{xiang2020detection,tang2021demon}. 

{More precisely, the objective of the attacker is to insert backdoors into a victim DNN $F_{\theta}$ and produce a backdoored DNN $F_{\theta}^{\mathrm{adv}}$,  which holds the following properties:
(1) $F_{\theta}^{\mathrm{adv}}$ misclassifies the \emph{source} samples patched with the attacker-selected trigger to the \emph{target} class, that is;
(2) $F_{\theta}^{\mathrm{adv}}$ maintains high test accuracy on clean test samples so as not to be detected by monitoring the test accuracy.
The backdoor is inserted by data poisoning, and the DNN is trained on the mixture of the triggered dataset (added by the attacker) $D_{\mathrm{adv}}$ and the genuine dataset $D_{\mathrm{train}}$, i.e., $D_{\mathrm{adv}} \cup D_{\mathrm{train}}$, generating the malicious model. 
As mentioned earlier, the attacker has full control of changing the data that are used for the training but has no direct access to manipulate the model parameters and the training process.}
%More precisely, the objective of the attacker is to insert backdoors into the model by data poisoning and make the output of the modified source inputs, i.e. the source inputs patched with the trigger, to be the attacker-chosen target. The DNNs are then trained on the mixture of poisoned dataset (added by the attacker) $D_{\mathrm{adv}}$ and the genuine dataset $D_{\mathrm{train}}$, i.e., $D_{\mathrm{adv}} \cup D_{\mathrm{train}}$, generating the malicious model. In this way, a backdoor is inserted into the malicious model so that the \emph{source} samples patched with the attacker-selected trigger will be misclassified to the \emph{target} class. At the same time, the malicious model maintains high test accuracy on clean test samples so as not to be detected by monitoring the test accuracy. As mentioned earlier, the attacker has full control of changing the data that are used for the training but has no direct access to manipulate the model parameters and the training process. 

\textbf{In this work, we consider the following attacks:} 
\begin{enumerate}
    \item \emph{Hidden trigger}: Hidden trigger backdoor attack~\cite{saha2020hidden} is one of the latest backdoor attacks. In this scenario, the attacker identifies poisoned images that are similar to the target in the image space but are close to the patched source, i.e., source images with triggers, in the representation space, and adds them to the genuine dataset. In this way, the poisoned images in the target class can bypass human inspection while the malicious model can still succeed in misclassifying the patched source images to the target label. The authors showed  that the conventional statistical anomaly detection methods failed to detect the poisoned data since their representations fuse with those of the genuine target data, exhibiting less separation. 

    \item \emph{TaCT}: It was first proposed in \cite{tang2021demon}, adding not only patched source images but also cover images, i.e., images from other classes attached with triggers, to the genuine dataset so as to weaken the dominance of the triggers and to obscure the difference between the poisoned data and target data. In this attack, 1$\%$ of the genuine images patched with the triggers are also added to the clean dataset as covering images.

    \item \emph{Badnets} \cite{gu2019badnets}: It is the conventional backdoor strategy where the patched source images with the target label are added to the genuine training dataset for manipulating the DNNs. 
\end{enumerate}

On the defender side, similar to the state-of-the-art defense techniques based on training data inspection, e.g., Activation Clustering \cite{chen2018detecting}, SCAn \cite{tang2021demon}, and Spectral Signatures \cite{tran2018spectral}, we assume that defenders have full access to the training data and the model under test so that they can collect the representations for each input sample, but they have no information about the poisoned data, for example the shape or position of the trigger. Furthermore, defenders have no control or visibility to the training process.

\section{Preliminaries} \label{preliminary}
In this section, we begin by defining and analyzing the representations that motivate our defense.

\textbf{Representations}: \emph{We denote the representations as the neuron activations learned by the layers of a neural network with given inputs. Formally, let $\mathbf{s}$ be an input sample and $f^l$ denote the mapping from $(l-1)$-th layer to $l$-th layer , then the representations $\mathbf{x}^L$ from the $L$th layer is defined as $\mathbf{x}^L=f^1(\cdots\ f^L(\mathbf{s}))$. For simplicity, we drop the superscript and denote $\mathbf{x}^L$ as $\mathbf{x}$.}

\textbf{Properties of representations}: A well-trained deep architecture can learn the underlying factors of variations determining the geometrical structure of raw data (inputs) \cite{cohen2020separability}. \emph{In this sense, data structures are flattened  across layers. Especially, at the last layer, representations lie on a Euclidean space, a.k.a. a linear space.}

\begin{table}[h!]
    \centering
    \caption{Flattening metric values for representations from different layers of a benign neural network. The metric values are getting smaller from input layers to the last layers, indicating the representations are flattened across layers. (This metric is derived from the difference between geodesic and Euclidean distances)}
    \vspace{-0.1in}
    %\small
    \begin{tabular}{c|c|c}
        \hline\hline
         \multirow{2}{*}{Layers} & \multicolumn{2}{c}{Flattening metric} \\
         \cline{2-3}
           & Stop sign & speed limit  \\
         \hline
        Input layer &0.2219 &0.3278  \\ 
        \hline
        Intermediate layer& 0.1551&0.2499   \\
        \hline
        Last layer &0.0411 &0.0831 \\
        \hline\hline
    \end{tabular}
    \label{metric}
    %\vspace{-0.1in}
\end{table}

To illustrate the above properties, 
we use a flattening metric which is  adapted from \cite{brahma2015deep} (please see Appendix \ref{ap4} for details). This metric is derived from the difference between geodesic and Euclidean distances. It should be equal to 0 when the data lie exactly in a linear space (since the geodesic and Euclidean distances are equal). We compute the metric using the representations from each layer of the respective DNNs. 
As an example, let us consider the GTSRB dataset and the corresponding  6Conv+2Dense network and present the flattening metric for the input layer, an intermediate layer, and the last layer in Table \ref{metric}. It can be observed that the value of the flattening metric decreases with increase in the depth of the network, and the values of the last layer are  small enough (close to 0), indicating representations from the last layer almost lie in a linear subspace. 
%Besides, from the previous work \cite{cohen2020separability}, deep neural networks transform representations from not linearly separable ones (in the first and intermediate layers) to linearly separable ones in the last layer. This verifies the above property, and thus, we draw the following remark:

\textbf{Remark 1.} It is reasonable to assume \emph{representations of higher/deeper layers (especially the last layer) for each object/class  approximately lie on a Euclidean subspace, i.e., a linear subspace.} 

This motivates us to focus on the representations of the last layers of the network, where processing becomes easier than on raw inputs, since, for example, linear techniques can be used.
\begin{table}[!t]
    \centering
    \caption{Flattening metric values for the representations from last layer of backdoored neural network. The accuracy of SVM indicates the linear separability of target and poisoned samples.}
    \vspace{-0.1in}
    %\small
    \begin{tabular}{c|c|c|c}
        \hline\hline         \multirow{2}{*}{(Target $\&$ poisoned)} & \multicolumn{2}{c|}{Flattening metric}& \multirow{2}{*}{SVM}\\
         \cline{2-3}
           & Target &Poisoned  \\
         \hline
              \begin{minipage}[b]{0.3\columnwidth}
		\centering
		\raisebox{-.5\height}{\includegraphics[scale=0.45]{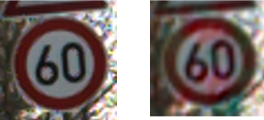}}
	\end{minipage} &0.0831 &0.0844&70.85  \\ 
        \hline
         \begin{minipage}[b]{0.3\columnwidth}
		\centering
		\raisebox{-.5\height}{\includegraphics[scale=0.45]{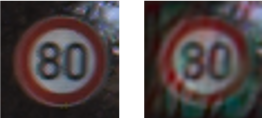}}
	\end{minipage} & 0.0978&0.1102& 77.18  \\
        \hline
             \begin{minipage}[b]{0.3\columnwidth}
		\centering
		\raisebox{-.5\height}{\includegraphics[scale=0.45]{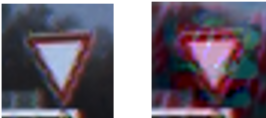}}
	\end{minipage} &0.0775 &0.0838 &68.18 \\
        \hline\hline
    \end{tabular}
    \label{metric_2}
    \vspace{-0.1in}
\end{table}

The signal (malicious activation) corresponding to triggers get boosted as we advance through the layers of the DNN \cite{tran2018spectral}. In other words, deeper layers manifest the trigger properties more conspicuously than earlier layers. Another line of research observed that the backdoored models tend to assign poisoned samples and benign samples to target classes based on different sets of features \cite{chen2018detecting}. For the benign samples, the output is based on the inherent features of the (true) target class; but for the poisoned samples, the final classification is influenced by the trigger features, which is different from the inherent features. Therefore, considering the infected target class, the samples (and their representations) belonging to this class can be viewed as a mixture of two groups: The poisoned samples and the benign samples \cite{tang2021demon}. In this case,  despite that the poisoned data and authentic data receive the same classification label, their mechanism of classification are different and this difference should be evident in the representations. Wang et al. \cite{wang2019neural}  also indicates that backdoored networks classify the poisoned data to the target class regardless of its original class mainly by changing the decision boundaries (instead of focusing on representation fusion). Besides the boundary change, the representations of the poisoned data can also fuse with those of the authentic data for some advanced attacks, e.g., dynamic trigger attacks and hidden trigger backdoor attacks, which makes it hard to group the representations into two distinct clusters and introduce more challenges for backdoor detection.
%
\begin{comment}
\begin{figure}[htbp!]
	\centering
    \subfigure[]{\includegraphics[scale=0.4]{{Imgs/ex1.png}}}
    %
    \subfigure[]{\includegraphics[scale=0.4]{{Imgs/ex2.png}}}
    \vspace{-0.2in}
    \caption{Representations of the last hidden layer projected onto the first 3 principle components. (a) representations of dog and cat (b) representations of  no entry and yield. The arrows represents the basis vectors. It is observed that the representations of authentic and poisoned data, where we can see they are approximately embedded in different spaces even they are partially overlapped.}
    \vspace{-0.1in}
    \label{ob}
\end{figure}
\end{comment}
\begin{figure}[!t]
	\centering
    \subfloat[]{\includegraphics[scale=0.5]{{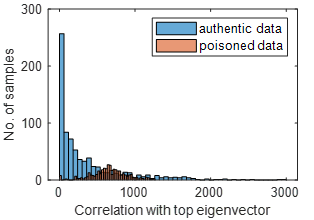}}}
    \hfil
    \subfloat[]{\includegraphics[scale=0.5]{{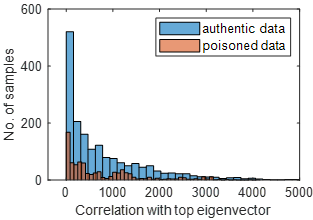}}}
    %\vspace{-0.1in}
    \caption{Plot of correlations for Spectral signatures \cite{tran2018spectral} of clean data and poisoned data from (a) ILSVARC2012 (b) GTSRB. We calculate the correlation of spectral signatures with the top eigenvector of covariance matrix of mixture representations.}
    \vspace{-0.2in}
    \label{corr}
\end{figure}
Furthermore, we observe that the representations of authentic data and poisoned data are still embedded in different linear spaces despite limited separation. To verify this observation, Table \ref{metric_2} summarizes the flattening metric values as well as the linear separability for three data pairs (e.g., each pair includes target class and poisoned class) in the last layer of backdoored network, where the linear separability is evaluated by the classification accuracy of support vector machine (SVM). We can see all their flattening metric values are small, indicating each class approximately lies in linear spaces. At the same time, classification accuracy based on SVM is not high, which implies that they cannot be clearly separated by linear techniques. Moreover, we present in Fig. \ref{corr} that the spectral signature \cite{tran2018spectral} of authentic representations and poisoned representations do not separate enough, since their correlations with the top eigenvector of the covariance matrix of representations are similar. This leads to our the second assumption:

\textbf{Remark 2.} \emph{The representations of authentic data and poisoned data from the last layer still lie approximately on different linear subspaces, which are not clearly separated.} 

%To this end, we propose a novel scheme which will be introduced in section \ref{methodology}. 

\noindent{\bf Data model:} Given a set of $m$ data points (without further declaration, we denote by data the  representations/activations from the last layer of the DNN) in $\mathbb{R}^n$, denoted by $\mathcal{X}$, which is the mixture of the targeted/authentic set and the poisoned/trigger set, we assume that
\begin{equation}
    \mathcal{X}=\mathcal{X}_1\cup\mathcal{X}_2
    \label{eq:1},
\end{equation}
where $\mathcal{X}_1$ denotes the targeted/authentic dataset and $\mathcal{X}_2$ is the poisoned/trigger dataset; $m_{\mathcal{1}}:=|\mathcal{X}_{1}|$ and $m_{\mathcal{2}}:=|\mathcal{X}_{2}|$. 

Our \textbf{data model} is then: 
\begin{equation}
    \mathbf{x}^{(\mathcal{c})}=\mathbf{z}^{(\mathcal{c})}+\mathbf{e},\ c=1,2
    \label{dm_vector},
\end{equation}
or
 \begin{equation}
    \mathbf{X}_{\mathcal{c}}=\mathbf{Z}_{\mathcal{c}}+\mathbf{E},\ c=1,2
    \label{dm_matrix},
\end{equation}
where $ \mathbf{X}_{\mathcal{c}}\in\mathbb{R}^{n\times m_{\mathcal{c}}}$  consists of $m_{\mathcal{c}}$ samples of  representations of dimensionality $n$ from the set $\mathcal{X}_{c}$, and $\mathbf{x}^{(\mathcal{c})}\in\mathbf{X}_{c}$ denotes a column of $\mathbf{X}_{c}$. The component $\mathbf{Z}_{\mathcal{c}}\in\mathbb{R}^{n\times m_{\mathcal{c}}}$  represents the object-specific representations (belonging to object $\mathcal{c}$) containing critical information for discrimination, and  $\mathbf{z}^{(\mathcal{c})}\in\mathbf{Z}_{c}$. On the contrary, $\mathbf{e}$ (without superscript) is a column of $\mathbf{E}\in\mathbb{R}^{n\times m_{\mathcal{c}}}$ containing common/global information shared by different objects. We can also denote $\mathbf{e}$ or $\mathbf{E}$ as noise/perturbation  that is useless  for classification. Furthermore, we designate the following conditions \emph{w.r.t.} the data model: 

\textbf{Condition 1}: \emph{Each object-specific representation, e.g., $\mathbf{z}^{(1)}$  or $\mathbf{z}^{(2)}$, perfectly lies in a linear subspace, e.g., $\mathcal{S}_{1}$ or $\mathcal{S}_{2}$, which has low rank structure; By contrast, each global representation $\mathbf{e}$ is independent identical distributed (i.i.d.) and is uncorrelated to object-specific ones, e.g., $\mathbf{z}^{(1)}$ and $\mathbf{z}^{(2)}$ .}

%\textbf{Condition 2}: \emph{The representation of each object/class, e.g., $\mathbf{x}^{(1)}$ or $\mathbf{x}^{(2)}$, is dominated by its object-specific component, e.g., $\mathbf{z}^{(1)}$ or $\mathbf{z}^{(2)}$, which is critical for discriminating different objects. On the contrary, the common component $\mathbf{e}$ which is useless for classification is suppressed. }

More specifically, \textbf{Condition 1} implies that any pair $\{\mathbf{e},\mathbf{z}^{(\mathcal{c})}\}$ satisfies   $\mathbf{e}^{\top}\mathbf{z}^{(\mathcal{c})}=0$, where the rationale is that if $\mathbf{e}^{\top}\mathbf{z}^{(\mathcal{c})}\neq0$ then $\mathbf{e}$ contains useful information for classification which can be incorporated into $\mathbf{z}^{(\mathcal{c})}$. 

%\textbf{Condition 2}  means $\mathbb{E}||\mathbf{e}||=\sigma\mathbb{E}||\mathbf{z}||$, where $\sigma<1$. The rationale behind is that the deep learning methods can build efficient classifiers where the representations of high layers are supposed to contain as less label-irrelevant information as possible to simplify the classifiers.

\section{Methodology}\label{methodology}
\subsection{Insight of the proposed algorithm}
\begin{comment}
In Section \ref{preliminary}, we know that the representations of poisoned and genuine inputs lie on two different linear subspaces, however, not separable enough by traditional linear techniques. To this end, we propose a coherence optimization based approach, where the coherence is defined as follows: 

\textbf{Definition 1} (Coherence). The coherence of a data point $\mathbf{x}\in\mathbb{R}^n$ with a space $\mathcal{S}$ spanned by $\mathbf{P}\in\mathbb{R}^{n\times k}$ is defined as:
\begin{equation}
    \mu=\|\mathbf{P}^\top\mathbf{x}\|
\end{equation}

The coherence above is upper bounded by $\|\mathbf{x}\|$ (which equals to 1 in our work since $\mathbf{x}$ is normalized to unit scale) when $\mathbf{x}$ lie in the space $\mathcal{S}$; it is lower bounded by 0 when $\mathbf{x}$ lie in the space $\mathcal{S}^\perp$ which is orthogonal to $\mathcal{S}$.
\end{comment}
Our proposed algorithm takes the advantage of the property that representations of authentic data and poisoned data approximately lie on different linear subspaces. We designate and solve a convex optimization problem which maximizes the coherence of weighted samples and a certain subspace, so as to obtain a sample-wise weight vector where its entries are well separated according to their associated subspace affiliations. The poisoned data can thus be detected and identified based on this weight vector, and finally mitigation of backdoor attacks can be realized. 

We give an example to illustrate the intuition of the proposed algorithm. For simplicity, we consider a noiseless case where representations perfectly lie in their associated subspaces. However, it should be noted that the conclusion drawn from noiseless case still holds for noisy case as perturbation/noise is uncorrelated with the object-specific components.
Consider two subspaces $\mathcal{S}_1$ and $\mathcal{S}_2$, and two associated datasets of representations, i.e., $\mathbf{X}_1$ and $\mathbf{X}_2$, where $\mathbf{X}_1$ denotes the authentic data and $\mathbf{X}_2$ denotes the poisoned data.  All samples are scaled to unit length, and all features are centralized using the information from clean testing data. Denote $\mathbf{P}_1$ and $\mathbf{P}_2$ the orthonormal basis matrix spanning $\mathcal{S}_1$ and $\mathcal{S}_2$, respectively. Let $\mathbf{X}=[\mathbf{X}_1, \mathbf{X}_2]$, and the sample-wise weight vector $\mathbf{a}^*$ be the optimal solution of the following optimization problem
\begin{equation}
    \begin{split}
 \underset{\mathbf{a}^\top\mathbf{a}=1}{\mathrm{max}}\ \mathbf{a}^\top\mathbf{X}^\top(\mathbf{I}-\mathbf{P}_1\mathbf{P}_1^\top)\mathbf{X}\mathbf{a}\\
    \end{split}.
    \label{Op-0}
\end{equation}
The optimization problem \eqref{Op-0} searches a unit weight vector that maximizes the coherence between weighted samples, i.e., $\mathbf{X}\mathbf{a}$, and a projected subspace, i.e., $\mathcal{S}_1^\perp\equiv\textrm{span}\{\mathbf{P}_1^{\perp}\}$ where $ (\mathbf{P}_1^{\perp})(\mathbf{P}_1^{\perp})^{\top} = \mathbf{I}-\mathbf{P}_1\mathbf{P}_1^{\top}$.  Since $\mathbf{X}_1$ is orthogonal to $\mathcal{S}_1^\perp$, any samples from $\mathbf{X}_1$ make no contribution to increasing the coherence between weighted samples and $\mathcal{S}_1^\perp$. Thus, the above optimization problem is more likely to assign small weights to $\mathbf{X}_1$ and give large weights to $\mathbf{X}_2$. To illustrate this, an extreme case is if all weights corresponding to $\mathbf{X}_1$  are non-zeros and others are zeros, the objective value of \eqref{Op-0} would be zero. In this way, there would exist a distinct gap between the weights associated with different datasets, which enables the detection and identification of poisoned dataset ($\mathcal{X}_1$)  by analyzing $\mathbf{a}^*$, e.g., using clustering techniques.

\begin{figure}[tbp!]
    \centering
    \includegraphics[scale=0.4]{{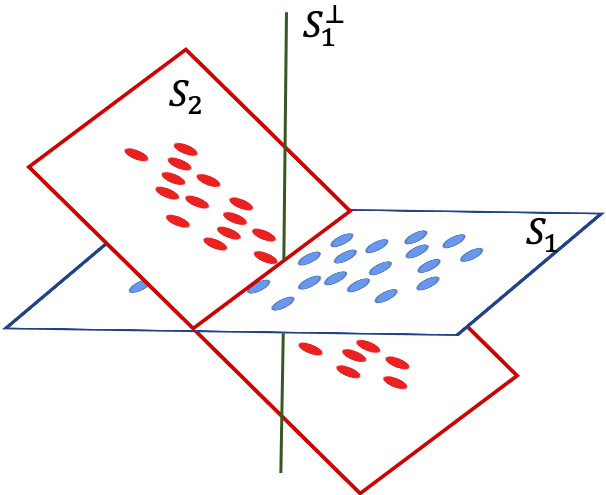}}
    \vspace{-0.1in}
    \caption{An intuitive example to illustrate the insight of our algorithm. The blue cycle denotes authentic representation and the red cycle denotes poisoned representations. $\mathcal{S}_1$ and $\mathcal{S}_2$ are 2-dimensional subspaces in a $\mathbb{R}^3$ space and the orthogonal subspace of $\mathcal{S}_1$ is a vector line. We can observe each point in $\mathcal{S}_1$ is orthogonal to $\mathcal{S}_1^\perp$ and thus makes no contribution at all to increase the coherence.}
%    \vspace{-0.3in}
    \label{ill}
\end{figure}

This intuition is illustrated in Fig. \ref{ill}, which shows two subspaces $\mathcal{S}_1$ and $\mathcal{S}_2$ along with a subspace $\mathcal{S}_1^\perp$ that is orthogonal to $\mathcal{S}_1$. If $\mathcal{S}_1$ and $\mathcal{S}_2$ are orthogonal to each other, then data points from $\mathcal{S}_1$ have zero coherence with the subspace $\mathcal{S}_2$.
% where $\mathcal{S}_1$ and $\mathcal{S}_2$ are 2-dimensional subspaces 
%
% in a $\mathbb{R}^3$ space and the orthogonal subspace of $\mathcal{S}_1$ is a vector line. We can observe each point in $\mathcal{S}_1$ is orthogonal to $\mathcal{S}_1^\perp$ and thus makes no contribution at all to increase the coherence.
\vspace{-0.1in}

%First, we target on expressing projection vectors (instead of self expressing data points) which are essentially the summaries of the data points when the data size is large enough. The advantage of taking use of projection vectors is they are more robust (as they are correlated with most of data points) and are more compact (as the rank of subspace is usually less than number of data points). Second, we minimize the the (subspace) coherence between projections and the linear combinations of data points to pointing out those data points from different subspace, instead of picking  up data points from the same subspace by fitting the projections by linear combinations of data points. Additionally, since datasets are mixed, the projections corresponding to each specific dataset are unknown, we use the estimated projections rather than the real projections. More precisely, we minimize the coherence between $\mathbf{P}_b$ ($\mathbf{P}_b\rightarrow\mathbf{P}_1)$ and the linear combinations of all data points ($\mathbf{X}^\top\mathbf{a}$, where $\mathbf{a}$ is the coefficient vector), and the coefficients associated to $\mathbf{X}_2$ are expected to be different from those with respect to $\mathbf{X}_2$. Therefore, the memberships of data points could be well indicated by their coefficients. 

\subsection{Problem Formulation and Optimization}
\label{ss:optimization}

%\textbf{Proposition 1}: \emph{The perturbations uncorrelated with object-specific components would not affect the coherence relations. More precisely,  
%let $\mathbf{x}_i$ and $\mathbf{x}_j$ be two data points in $\mathcal{X}$.  Then, $\mathbb{E}\{||\mathbf{x}^\top_i\mathbf{x}_j||^2\}=\mathbb{E}\{||\mathbf{z}^\top_i\mathbf{z}_j||^2\}+C$, where $C$ is a constant. }

We then extend the optimization problem in Eq.\eqref{Op-0} to a more generalized case, where the ground truth labels (e.g., which class they originally come from) of $\mathbf{X}_1$ and $\mathbf{X}_2$ are unknown. We estimate a projection matrix $\mathbf{P}$ which is expected to be closer to $\mathbf{P}_1$ than $\mathbf{P}_2$, so that a sample from $\mathbf{X}_1$ contributes less to the objective function than a sample from $\mathbf{X}_2$. Correspondingly, it is more likely to assign larger weights to $\mathbf{X}_2$ than $\mathbf{X}_1$, and thus there would be a  gap in the weights of the sample-wise weight vector $\mathbf{a}^*$.  Given a proper $\mathbf{P}$, we formulate the following optimization problem
\begin{equation}
 \underset{\mathbf{a}^\top\mathbf{a}=1}{\mathrm{max}}\ \mathbf{a}^\top\mathbf{X}^\top(\mathbf{I}-\mathbf{P}\mathbf{P}^\top)\mathbf{X}\mathbf{a},
    \label{Op-1}
\end{equation}
where $\mathbf{X}=[\mathbf{X}_1, \mathbf{X}_2]$, $\mathbf{X}_1$ denotes the authentic data and $\mathbf{X}_2$ denotes the poisoned data. All samples are scaled to unit length, and all features are centralized by the information from clean testing data. 

As Eq.\eqref{Op-1} is in fact an eigenvalue decomposition problem which is tractable,
the optimal $\mathbf{a}^*$ could be directly obtained by performing eigenvalue decomposition of $\mathbf{X}^\top(\mathbf{I}-\mathbf{P}\mathbf{P}^\top)\mathbf{X}$,  and $\mathbf{a}^*$ is the eigenvector corresponding to the largest eigenvalue. 
It is noted that $\mathbf{P}$ plays an important role in above optimization problem, More specifically, we demonstrate in our experiments that \emph{the smaller the angle between $\mathcal{S}\equiv\mathrm{span}\{\mathbf{P}\}$ and $\mathcal{S}_1$, the larger the gap between the weights w.r.t. $\mathbf{X}_1$ and  $\mathbf{X}_2$.} Then, we propose a simple way to calculate $\mathbf{P}$ in the backdoor attack scenario where  the variations of poisoned representations are more likely to be smaller than those of the authentic ones in the target class (this may be attributed to the sample size of poisoned samples as well as the stealthy objective of backdoor attack). We perform an eigenvalue decomposition on the mixture representations in the target class and select the first $k$ eigenvectors as $\mathbf{P}$. %In this way, it is easy to derive that $\mathcal{S}$ is closer to $\mathcal{S}_1$ than $\mathcal{S}_2$. %Besides, our extensive experimental results in Section \ref{} showcase that $\mathbf{P}$ obtained in this way works well.
%\vspace{-0.1in}

\subsection{Detection and Mitigation based on Sample-wise Weight Vector}
\begin{comment}
Before diving into the details of our algorithm, we have the following proposition that enables the detection.

\textbf{Proposition 1}. \emph{If the optimal $\mathbf{a}^*$ approximately follows a single normal distribution, the representations are not contaminated, i.e., $\mathcal{X}=\mathcal{X}_1$. By the contrast, if the optimal $\mathbf{a}^*$  does not follow a  normal distribution, then the representations are  contaminated, i.e., $\mathcal{X}=\mathcal{X}_1\cup\mathcal{X}_2$.}

\textbf{Note:} We provide detailed analysis to show how above proposition holds in Appendix \ref{Ap1}.  
\end{comment}
The intuitive idea behind detection is that, the elements of $\mathbf{a}^*$ would be bi-modal if the representations are contaminated, which is because of the aforementioned gap in the weights of samples from two groups, the genuine and the poisoned; on the contrary, the elements of $\mathbf{a}^*$ would be uni-modal if the representations are not contaminated.
%\textbf{Proposition 1} is natural, since for the well-trained neural network, the representations from one class should be uni-modal and those from infected class are bi-modal. 
%\textbf{Proposition 1} is natural, since for the well-trained neural network, the representations from one class should be uni-modal and those from infected class are bi-modal.

To this end, we adopt likelihood ratio test \cite{fan2001generalized} to judge whether the representations are contaminated or not. If a uni-modal distribution such as a single Gaussian fits $\mathbf{a}^*$ better, the corresponding class is authentic; if a bi-modal distribution like the mixture of two Gaussian fits $\mathbf{a}^*$ better, the corresponding class is infected. More precisely, for each class $t$, we test the following null hypothesis,

$\mathbf{H}_0$: $\mathbf{a}^*$ is drawn from a Gaussian distribution;
%
%Against the alternative one (considering representations in target class are essentially mixed with data from two classes, we assume they follow a two-component mixture normal distribution),

$\mathbf{H}_1$: $\mathbf{a}^*$ is drawn from a two-component Gaussian mixture distribution.
    
The test is based on the following statistic:
\begin{equation}
\begin{split}
\mathit{J_t}=-2\mathrm{log}\frac{\mathcal{L}(\mathbf{a}^*|\mathbf{H}_0)}{\mathcal{L}(\mathbf{a}^*|\mathbf{H}_1)}
\end{split}
\label{Jt}
\end{equation}
where $\mathcal{L}(\mathbf{a}^*|\mathbf{H}_0)$ denotes the estimated likelihood of $\mathbf{a}^*$ under the null hypothesis, and $\mathcal{L}(\mathbf{a}^*|\mathbf{H}_1)$ represents the likelihood of $\mathbf{a}^*$ under the alternative hypothesis. It should be noting that  the likelihood functions are estimated by EM algorithm \cite{moon1996expectation}. 

The $J_t$ statistic follows a \textit{chi-square} distribution with the degrees of freedom being equivalent to the difference of free parameters between the two hypothesis \cite{fan2001generalized}. Since $J_t$ is not symmetrically distributed, we use the technique based on a robust scale estimator, \emph{Absolute Pairwise Difference (APD)} \cite{rousseeuw1993alternatives}, to detect the class with great values of $J_t$. More precisely, we further compute the following statistic, which is termed as \emph{anomaly index}, for each class $t$
 \begin{equation}
     \begin{split}
         &\widehat{J}_t=\frac{|J_t-\mathrm{med}_t(J_t)|}{c\ APD(J_t)}\\
     \end{split}
     \label{Jt1}
 \end{equation}
 where $\mathrm{med}$ denotes the sample median and $c=1.1926$ is the correct factor making \emph{APD} unbiased towards finite samples. In this way, we flag those $J_t$ as spurious for which $\widehat{J}_t$ exceeds a threshold ($\tau$) with confidence level  $\alpha$. Furthermore, we focus on those classes with $J_t>\mathrm{med}_t(J_t)$.
 The specific procedures of the detection algorithm are summarized in Algorithm \ref{A:1}.
\begin{algorithm}[tbp] 
 \caption{Detection of trigger dataset} 
 \label{A:1}
 \begin{algorithmic}[1]
\Require The coefficient vector $\mathbf{a}^*$ for each class $t\in \mathcal{T}\equiv\{1,\cdots,T\}$, and a empty set $\tilde{\mathcal{T}}\equiv\phi$; 

\State Compute the statistic $\widehat{J}_t$ as indicated in Eq.\eqref{Jt} and Eq.\eqref{Jt1};

\State $\tilde{\mathcal{T}}\leftarrow \{t:\widehat{J}_t>\tau,\ J_t>\bar{J}\}$; 

\Ensure The infected class $t\in\tilde{\mathcal{T}}$.
\end{algorithmic}
\end{algorithm}
%\vspace{-0.1in}

In the following stage, we identify those poisoned data points and remove them to achieve backdoor attack mitigation.
Our identification method is derived based on the following property,

\textbf{Remark 3}. \emph{The optimized weights have a grouping effect. More precisely, let $\mathbf{a}^*$ be the solution of Eq.\eqref{Op-1}, we have 
\begin{equation}
    |\mathit{a}^*_i-\mathit{a}^*_j|\leq\sqrt{\frac{2(1-\rho_{ij})}{\lambda^*}}
\end{equation}
where $\lambda^*$ is the square of the maximum eigenvalue of $\mathbf{X}^\top(\mathbf{I}-\mathbf{P}\mathbf{P}^\top)\mathbf{X}$,  $\mathit{a}^*_i$ and $\mathit{a}^*_j$ are elements of $\mathbf{a}^*$, and $\rho_{ij}=\mathbf{x}^\top_i\mathbf{x}_j$ denotes the coherence between samples.} 
See Appendix \ref{Ap2} for more details.

Since $\sqrt{\lambda^*}$ is the largest eigenvalue of $\mathbf{X}^\top(\mathbf{I}-\mathbf{P}\mathbf{P}^\top)\mathbf{X}$, which is much larger than 1 in practice (as derived from different classes from ILSVRC2012 and GTSRB datasets, $\lambda^{*}$ ranges from 15 to 30), it is easy to observe that the difference between $\mathit{a}^*_i$ and $\mathit{a}^*_j$ is extremely small (close to 0) when  $\mathbf{x}_i$ and $\mathbf{x}_j$ are highly correlated ($\rho_{ij}$ is close to 1).
\begin{comment}
\begin{figure}[h!]
\centering
\includegraphics[scale=0.55]{{Imgs/lambda.png}}
\caption{$\lambda^*$ for different datasets adapted from GTSRB and ILSVRC2012.}
\label{lambda}
\end{figure}
\end{comment}

\textbf{Remark 3} provides a theoretical guarantee that highly correlated data points can be grouped into the same cluster by analyzing the weight vector. Again, we have demonstrated that there is distinct gap among the entries of the weight vector. These jointly implies, \emph{the identification of poisoned samples could thus be achieved by using the state-of-the-art clustering algorithms.} In this work, the simple clustering algorithm, i.e., "k-means" \cite{teknomo2006k} can well separate the poisoned samples from the authentic ones based on our optimized weight vector, which also highlight the effectiveness of the proposed coherence optimization algorithm. We summarize the identification and mitigation procedure in Algorithm 2.
\vspace{-0.1in}
%of trigger data points, we optimize the following problem \cite{}
%\begin{equation}
%    \begin{split}
%        \underset{\mathbf{w}}{\mathrm{min}}&\ ||\mathbf{w}||^2+\lambda||\mathbf{w}||_0\\
%        \mathrm{s.t.} &\ \mathbf{X}^\top\mathbf{a}=\mathbf{X}^\top\mathbf{w}
%    \end{split},
%    \label{Op-sparse}
%\end{equation}
%where the $l_0$-norm penalty encourages the sparsity of $\mathbf{w}$ and the $l_2$-norm regularization can be regarded as the consideration of data correlations due to its strictly convex property\cite{}. However, the optimization problem above is NP-hard due to the $l_0$-norm penalty, we naturally relax it as follows
%\begin{equation}
    %\begin{split}
        %\underset{\mathbf{w}}{\mathrm{min}}&\ ||\mathbf{X}^\top\mathbf{a}-\mathbf{X}^\top\mathbf{w}||^2+\lambda_1||\mathbf{w}||^2+\lambda_2||\mathbf{w}||_1\\
%    \end{split},
%    \label{Op-2-relax}
%\end{equation}
%which is an elasticnet problem \cite{} and can be effectively solved by the LARS-EN algorithm \cite{}. By properly solving the above optimization problem, we could obtain a sparse vector $\mathbf{w}$ whose non-zero entries correspond to the data points that have smaller coherence with $\mathbf{X}^\top\mathbf{a}$, i.e., trigger/poisoned data points. The segmentation algorithm is provided in Algorithm \ref{A:2} in what follows,
\begin{algorithm}[tbp] 
 \caption{Mitigation of the backdoor attack} 
 \label{A:2}
 \begin{algorithmic}[1]
\Require The mixture representation  $\mathbf{x}_i$, where $i\in\mathcal{I}\equiv\{1,\cdots,m\}$ and weight vector $\mathbf{a}=[a^*_1,\cdots,a^*_m]^\top$;  

\Ensure  Retrained deep neural network model using the `cleaned' data. 

\State Implement "k-means" algorithm to divide $\mathbf{a}^*$ into two clusters (steps 2-6);

\State \textbf{Initialize:} Set cluster number $k\leftarrow2$, iteration index $t\leftarrow0$ and $\mathcal{I}^{[0]}_1\leftarrow\phi,\mathcal{I}^{[0]}_2\leftarrow\phi$, and  randomly assign two centres $c_1^{[0]}$ and $c_2^{[0]}$;

\While {not converge}

\State Assign each element of $\mathbf{a}^*$ , i.e., $a^*_p$, to the cluster/set ($\mathcal{I}^{[t]}_i$) with the nearest center ($c^{[t]}_i$):
\begin{equation}
    \mathcal{I}^{[t]}_i=\{p:\|\mathit{a}^*_p-c^{[t]}_i\|^2\leq\|\mathit{a}^*_p-c^{[t]}_j\|^2, j=1,2\}
\end{equation}

\State Recalculate centroids for observations assigned to each cluster.
\begin{equation}
    c^{[t+1]}_i=\frac{1}{|\mathcal{I}^{[t]}_i|}\sum_{a^*_p\in\mathcal{I}^{[t]}_i}a^*_p
\end{equation}

\EndWhile

\State $\tilde{\mathcal{I}}\leftarrow\{\mathcal{I}_i:|\mathcal{I}_i|\geq|\mathcal{I}_j|,j=1,2\}$ and $\bar{\mathcal{I}}\leftarrow\mathcal{I}_{\cup}\tilde{\mathcal{I}}$. The "cleaned" representation $\tilde{\mathbf{x}}_i$ satisfies $i\in\tilde{\mathcal{I}}$ and the poisoned representations $\bar{\mathbf{x}}_i$ satisfies $i\in\bar{\mathcal{I}}$.

\end{algorithmic}
\end{algorithm}
%\vspace{-0.2in}

\subsection{Computational Complexity}
Since the proposed coherence optimization problem in Eq.\eqref{Op-1} can be solved by a simple eigenvalue decomposition on $\mathbf{X}^\top(\mathbf{I}-\mathbf{P}\mathbf{P}^\top)\mathbf{X}\in\mathbb{R}^{n\times n}$, its computational complexity is $O(n^3)$. For the detection algorithm, the major computational burden is produced by the EM algorithm that is used to estimate the likelihood function in each hypothesis. The complexity of EM algorithm is generally $O(mt)$, where $m$ is the sample size and $t$ is the iteration steps. For the mitigation algorithm, the computational complexity of "k-means" is typically $O(m^3)$. The overall complexity of our method is $O(mt+m^3+n^3)$.

\section{Experimental Results}
\label{exp}
\begin{figure}[t]
	\centering
    \subfloat{\includegraphics[scale=0.25]{{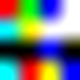}}}
    \hfil
    \subfloat{\includegraphics[scale=0.25]{{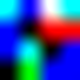}}}
    \hfil
    \subfloat{\includegraphics[scale=0.25]{{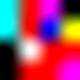}}}
    \hfil
    \subfloat{\includegraphics[scale=0.25]{{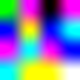}}}
    \hfil
    \subfloat{\includegraphics[scale=0.25]{{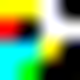}}}
    \hfil
    \subfloat{\includegraphics[scale=0.25]{{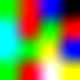}}}
    \hfil
    \subfloat{\includegraphics[scale=0.25]{{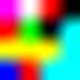}}}
    \hfil
    \subfloat{\includegraphics[scale=0.25]{{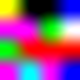}}}
    \hfil
    \subfloat{\includegraphics[scale=0.25]{{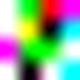}}}
    \hfil
    \subfloat{\includegraphics[scale=0.25]{{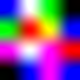}}}
%    \vspace{-0.2in}
    \caption{The square triggers used in the experiments.}
    \label{triggers}
    \vspace{-0.1in}
\end{figure}

In this section, we evaluate our backdoor detection scheme on the ILSVRC2012 dataset \cite{russakovsky2015imagenet} (built for recognizing general objects) and GTSRB dataset \cite{stallkamp2012man} (built for recognizing traffic signs) for three attack schemes: (1) Hidden trigger \cite{saha2020hidden}; (2) TaCT \cite{tang2021demon}; and (3) Badnets \cite{gu2019badnets}. We use  AlexNet~\cite{krizhevsky2012imagenet} for the ILSVRC2012 dataset, and a simplified network of 6 convolution layers and 2 dense layers (see Table~\ref{architecture for GTSRB} in the Appendix) for the GTSRB dataset. For each infected model, we use the randomly generated square triggers in \cite{saha2020hidden} as shown in Fig~\ref{triggers}. For the ILSVRC2012 dataset, the trigger is of size $30 \times 30$ with \emph{varying positions}. Since the input size of the GTSRB dataset is only $32 \times 32$, we set the trigger size as $8 \times 8$ and fix it at the bottom right corner of the images. The adopted experimental method follows well-established evaluation methods in the state-of-the-art \cite{saha2020hidden,wang2019neural,tang2021demon}.
\vspace{-0.1in}
\subsection{Infected models}
%\textbf{Hidden trigger}: In Hidden trigger attack, poisoned images, which look like the target in the image space but are close the patched source in the representation space, are learned and added to the clean training dataset. In this manner, the poisoned images can bypass the human inspection and the malicious model can still succeed in misleading the patched source images to the target label. The authors showed in their paper that the conventional statistic anomaly detection methods failed to detect the poisoned data as they don't have much separation with the target data. 
%
%\textbf{TaCT}: TaCT was first proposed in [], which adds not only patched source images but also the covering images (images from other classes attached with triggers) to the clean dataset so as to weaken the dominance of the trigger-carrying images and obscures the difference between the patched source images and target images. Following the strategy in [], 1$\%$ of the clean images patched with the triggers are also added to the clean dataset as covering images.
%
%\textbf{Badnets}: Badnets is the conventional backdoor strategy where the source images attached with triggers (patched source images) with the target label are added to the clean training dataset for manipulating the deep neural networks. 
%
\subsubsection{Random ILSVRC2012 source-target pairs}
\label{ILS pairs}
We choose 10 random pairs of source and target categories from ILSVRC2012 to evaluate our defense strategy. We use 800 genuine images for each class and add 400 poisoned samples to the target class, similar to hidden trigger attack setup~\cite{saha2020hidden}. The classification accuracy (CA) and the attack success rates (ASRs) are summarized in Table~\ref{infected models}. In summary, the infected models can achieve comparable classification accuracy as the clean models and the ASRs are above 70\% for hidden trigger attacks and around 90\% for the TaCT and badnets. %Especially, the ASRs increase as the number of poisoned samples increase. 
\begin{comment}
\begin{table}[]
    \centering
    \begin{tabular}{c|c|c|c|c|c|c|c|c|c}
        Pair & \multicolumn{2}{c}{1} & \multicolumn{2}{c}{2} & \multicolumn{2}{c}{3} \\
        \multirow{2}{*}{Badnets} & CA & ASR \\ & 97\% & 70\% \\
        \multirow{2}{*}{Hidden trigger} & CA & ASR \\ & 97\% & 80\% \\
        \multirow{2}{*}{Badnets} & CA & ASR \\  & 97\% & 80\% 
    \end{tabular}
    \caption{Caption}
    \label{tab:my_label}
\end{table}
\end{comment}
%
\subsubsection{Selected GTSRB source-target pairs}
\label{GTSRB}
We also evaluate our method on the GTSRB dataset for showing how our method defends against backdoor attacks in traffic sign identification models. We choose three source-target pairs which will cause life-risking threats in real world. The GTRSB pairs are listed in Table~\ref{pairs for GTSRB}. We add 800 poisoned samples for each source-target pairs. It is noted that we also pick two speed limit signs (speed limit of 30 and 80) with similar features, so that we can evaluate how our defense works for similar target and source classes. The performance of the clean and infected models are summarized in Table~\ref{infected models}. In general, the infected models hold high classification accuracy and high attack success rate. %
%
\begin{comment}
\begin{table*}[thpb]
    \centering
    \begin{tabular}{c|c|c|c|c|c|c}
        \hline
         \multirow{3}{*}{ } & \multicolumn{3}{c|}{ILSVARC2012} & \multicolumn{3}{c}{GTSRB} \\
         \cline{2-7}
          & clean & \multicolumn{2}{c|}{Infected} & clean & \multicolumn{2}{c}{Infected} \\
         \cline{2-7}
           & CA & CA & ASR & CA & CA & ASR \\
         \hline
        Hidden trigger & 92.1$\pm$1.30\% & 91.5$\pm$1.36\% & 65.2$\pm$12.6\% & 99.5\% & 99.0\% & 80\% \\
        \hline
        \tabincell{c}{TaCT} & 95.1\% & 95.0\% & 85.2$\pm$2.37\% & 99.5\% & 99.1\% & 80\% \\
        \hline
        Badnets & 95.1\% & 94.8\% & 86.3$\pm$1.50\% & 99.5\% & 99.5\% & 84\% \\
        \hline
    \end{tabular}
    \caption{The performance of clean and infected models}
    \label{infected models}
\end{table*}
\end{comment}
%
\begin{table}[t]
    \centering
    \caption{Selected source-target pairs for GTSRB}
    \vspace{-0.1in}
    %\small
    \begin{tabular}{c|c}
    \hline
    \hline
     \textbf{source} & \textbf{target} \\
    \cline{1-2}
      \begin{minipage}[b]{0.3\columnwidth}
		\centering
		\raisebox{-.5\height}{\includegraphics[scale=0.15]{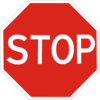}}
	\end{minipage}
   &       \begin{minipage}[b]{0.3\columnwidth}
		\centering
		\raisebox{-.5\height}{\includegraphics[scale=0.15]{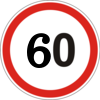}}
	\end{minipage} \\
	stop sign & speed limit of 60 \\
    \cline{1-2}
      \begin{minipage}[b]{0.3\columnwidth}
		\centering
		\raisebox{-.5\height}{\includegraphics[scale=0.15]{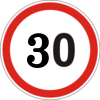}}
	\end{minipage}
   &       \begin{minipage}[b]{0.3\columnwidth}
		\centering
		\raisebox{-.5\height}{\includegraphics[scale=0.15]{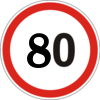}}
	\end{minipage} \\
      speed limit of 30 & speed limit of 80 \\
    \cline{1-2}
      \begin{minipage}[b]{0.3\columnwidth}
		\centering
		\raisebox{-.5\height}{\includegraphics[scale=0.15]{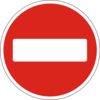}}
	\end{minipage}
   &       \begin{minipage}[b]{0.3\columnwidth}
		\centering
		\raisebox{-.5\height}{\includegraphics[scale=0.15]{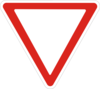}}
	\end{minipage} \\
      no entry & yield \\
    \hline
    \hline
    \end{tabular}
    \vspace{-0.1in}
    \label{pairs for GTSRB}
\end{table}
\begin{table}[t]
    \centering
    \caption{The performance of clean and infected models}
    \label{infected models}
    \vspace{-0.1in}
    \footnotesize
    \begin{tabular}{c|c|c|c|c|c|c}
        \hline\hline
         \multirow{3}{*}{Attack Type} & \multicolumn{3}{c|}{ILSVARC2012} & \multicolumn{3}{c}{GTSRB} \\
         \cline{2-7}
          & clean & \multicolumn{2}{c|}{Infected} & clean & \multicolumn{2}{c}{Infected} \\
         \cline{2-7}
           & CA & CA & ASR & CA & CA & ASR \\
         \hline
        Hidden trigger & 93.5\% & 92.7\% & 78.2\% & 96.7\% & 96.0\% & 84.1\% \\
        \hline
        \tabincell{c}{TaCT} & 93.5\% & 92.7\% & 88.7\% & 96.7\% & 96.1\% & 96.4\% \\
        \hline
        Badnets & 93.5\% & 93.0\% & 90.5\% & 96.7\% & 96.5\% & 96.5\% \\
        \hline\hline
    \end{tabular}
    \vspace{-0.2in}
\end{table}
\subsection{Performance of the proposed defense}
In this section, we specify the steps of implementing our defense as described in Section~\ref{methodology}, including identifying the infected class based on the optimized weights and locating the poisoned samples by clustering the weights. The infected models can be repaired by our method by removing the poisoned samples and retraining the model from scratch or relabeling the poisoned samples with their source class and continuing the training. Experiments are performed on a 2.9 GHz Intel i7 processor with 32 GB of RAM.

\subsubsection{Infected class detection via optimized sample weights}
\begin{figure*}[th!]
	\centering
    \subfloat[]{\includegraphics[scale=0.65]{{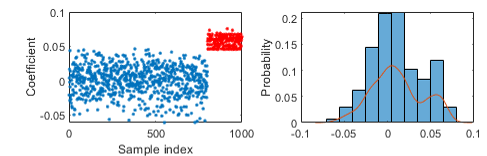}}}
    \hfil
    \subfloat[]{\includegraphics[scale=0.65]{{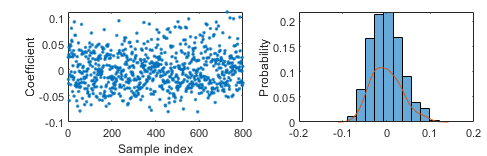}}}
%    \vspace{-0.2in}
    \caption{The optimized coefficients for (a) the infected class and (b) the clean class. The distribution of coefficients for the infected class follows a two Gaussian mixture, while distribution of coefficients for the clean class follows single Gaussian distribution.}
    %\vspace{-0.15in}
    \label{f:coefficient}
\end{figure*}

\begin{comment}
\begin{figure*}[h!]
	\centering
    \subfigure[]{\includegraphics[scale=0.63]{{Imgs/sj3.png}}}
    \subfigure[]{\includegraphics[scale=0.63]{{Imgs/sj2.png}}}
    \subfigure[]{\includegraphics[scale=0.63]{{Imgs/sj1.png}}}
    \vspace{-0.2in}
    \caption{The scaled likelihood ratio index for infected class and clean class upon different GTSRB source-target pairs (a,b,c). The scaled likelihood ratio is computed via dividing the likelihood ratio index by its corresponding clean class value. In this way, all clean class indexes are transformed to 1 so that the difference between infected class and clean class are highlighted. The blue bar denotes the index based on optimized coefficient (by our method); the green bar denotes the index upon top ten features (as AC \cite{chen2018detecting} does) and the yellow bar is upon the original neuron representations (as SCAn does). }
    \vspace{-0.1in}
    \label{f:sj} 
\end{figure*}
\end{comment}

First, we evaluate how accurately our defense can identify the infected classes. \textit{In our methodology, we aim to distinguish the representations of the poisoned samples from those of the clean samples, which can significantly overlap (see Fig~\ref{f:o1}), 
%(please see an example in Fig~\ref{f:o1}, and refer to Appendix for more examples),
and thus are hard to separate into two corresponding groups by conventional clustering methods, like KNN and K-means.}

Based on Eq.~\eqref{Op-1}, the optimized sample weight vector can be calculated based on the learned representations for each class. For the infected class, it is obvious that the weights for the authentic and poisoned samples form two distinguishable clusters, while the weights for the clean class almost follow the Gaussian distribution. The optimized weights for the infected and clean class are shown in Fig~\ref{f:coefficient}, from which we can see different distributions of the optimized weights in infected class and clean class. 
%In Fig~\ref{f:sj}, we present the scaled likelihood ratio for infected class and its associated clean class to showcase the distinguishability of the optimized weights in our method, where the scaled likelihood ratio is calculated via dividing the likelihood ratio index in \eqref{Jt1} by its associated clean value, e.g., $J_t/J_{\mathrm{clean}}$. We can observe that the scaled likelihood ratio of infected class goes away from that of the clean class, indicating that our method can successfully detect the attacks.
%For comparison, we also compute the same index based on the features/representations that are used for detection in AC \cite{chen2018detecting} and SCAn. It is noted that our method enjoys the largest value of scaled likelihood ratio for infected class, which means our method has the largest probability to accept the $\mathbf{H}_1$ hypothesis, \emph{a.k.a.}, data is contaminated. To this end, our method demonstrates superior performance upon detecting backdoor attacks.

\begin{figure}[t]
	\centering
    \subfloat[]{\includegraphics[scale=0.48]{{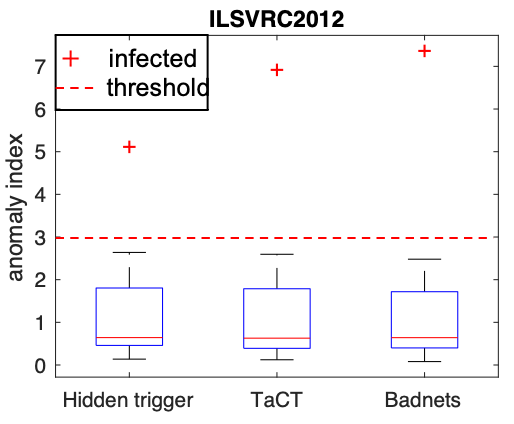}}}\
    \subfloat[]{\includegraphics[scale=0.48]{{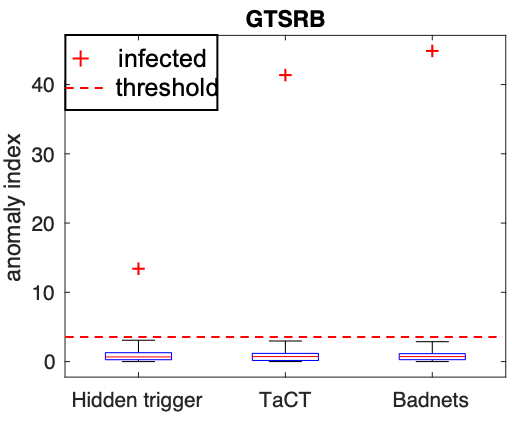}}}
%    \vspace{-0.2in}
    \caption{The anomaly index for (a) the ILSVARC2012 and (b) the GTSRB dataset. The red dash line is the threshold 3 with 99.9\% confidence level.}
    \vspace{-0.2in}
    \label{f:ai1}
\end{figure}

%For the clean classes, the coefficients for each class are estimated by both one Gaussian distribution.
Then, for each class, we fit the distribution of optimized weights via the mixture of Gaussian distribution(s) with different number of kernel(s) and report the likelihood ratio along with the anomaly index defined in Eq.~\eqref{Jt1}. The anomaly index for one class exceeding a threshold ($\tau$) indicates that this class is infected with corresponding confidence level. As suggested by \cite{leys2013detecting}, we vary the threshold ($\tau$) using the values [2, 2.5, 3], which means the confidence level of the optimized weight vector drawn from a two-component of mixture normal distribution --that is, the corresponding class is infected-- varies from $97.7\%$ to $99.9\%$. More specifically, \cite{leys2013detecting} suggests a threshold of 2.5 as a conservative choice for univariable statistical problems, with 3 being "very conservative" and 2 "poorly conservative" choices respectively.
%we set the threshold of anomaly index to be 3 (which indicates the level of confidence of the detection results is $99\%$). Accordingly, the class with anomaly index exceed 3 and likelihood ratio larger than their median is reported as the infected class. 
The detection rates are summarized in Table \ref{detection rates}. We observe our PiDAn algorithm can achieve superior backdoor detection performance with infected class detection rates around 96\% and less than 5.5\% clean classes being misclassified as the infected.
%We observe that our method achieves comparable overall detection performance with the recent detection approach, e.g., SCAn. 
%Moreover, our method outperforms SCAn in lower false positive rates for the more advanced attack (hidden trigger). It is noted that the threshold for the anomaly index is set to be 7 in SCAn. 
Shown in Fig~\ref{f:ai1}, the anomaly index of infected classes all exceed 3, but the indexes for the hidden trigger attack are much smaller than those of the other two attacks, which indicates the representations of poisoned and clean data for hidden trigger attack are less separable than the others and it is harder to be detected.
%\vspace{-0.1in}
\begin{table*}[htpb]
    \centering
    \caption{The detection performance on two datasets (ILSVARC2012 dataset and GTSRB) and three backdoor schemes (Hidden trigger, TaCT and Badnets) between different methods (Activation clustering \cite{chen2018detecting}, SCAn \cite{tang2021demon} and PiDAn). $\mathrm{TPR}$ is the rate of detecting the infected class and $\mathrm{FPR}$ is the rate of detecting clean class as the infected (Activation clustering set different thresholds for the Silhouette Score in their method of detecting the infected class and "AC-x" means the class will be detected as infected when the corresponding Silhouette Score is larger than "x"; "PiDAn-x" indicates that the class will be detected as infected when the corresponding anomaly index we defined in Eq.~\eqref{Jt1} is larger than "x").}
    \vspace{-0.1in}
    %\small
    \begin{tabular}{c|c|c|c|c|c|c|c|c|c|c|c|c}
        \hline\hline
         \multirow{3}{*}{\tabincell{c}{Detection \\Method}} & \multicolumn{6}{c|}{ILSVARC2012} &\multicolumn{6}{c}{GTSRB}  \\
          \cline{2-13}
         &\multicolumn{2}{c|}{Hidden trigger}&\multicolumn{2}{c|}{TaCT}&\multicolumn{2}{c|}{Badnets} &\multicolumn{2}{c|}{Hidden trigger}&\multicolumn{2}{c|}{TaCT}&\multicolumn{2}{c}{Badnets}\\
         \cline{2-13}
            & TPR & FPR  & TPR & FPR & TPR & FPR & TPR & FPR  & TPR & FPR & TPR & FPR\\
         \hline
%        Activation clustering & 91.5\% & 65.2\% & 99.0\% & 80 \% &91.5\% & 65.2\% & 99.0\% & 80 \%  &91.5\% & 65.2\% & 99.0\% & 80\\
%        \hline
        AC-0.10 & 50.0\% & 39.5\% & 50.0\% & 38.4\% & 51.0\% & 38.4\% & 76.6\% & 60.2\% & 100.0\% & 62.4\% & 100.0\% & 59.5\% \\
        \hline

        AC-0.11 & 10.0\% & 32.6\% & 46.0\% & 31.6\% & 48.0\% & 31.1\% & 60.0\% & 47.1\% & 66.7\% & 45.5\% & 66.7\% & 44.8\% \\
        \hline

        AC-0.12 & 5.0\% & 20.1\% & 10.0\% & 21.6\% & 8.0\% & 19.5\% & 33.3\% & 34.8\% & 63.3\% & 35.5\% & 66.7\% & 32.4\% \\
        \hline

        AC-0.13 & 4.0\% & 14.2\% & 5.0\% & 16.3\% & 5.0\% & 15.8\% & 30.0\% & 34.8\% & 53.3\% & 33.1\% & 60.0\% & 30.1\% \\
        \hline
        
        AC-0.14 & 0.0\% & 6.8\% & 5.0\% & 6.3\% & 5.0\% & 5.3\% & 16.7\% & 25.2\% & 16.7\% & 23.6\% & 26.7\% & 24.5\% \\
        \hline

        AC-0.15 & 0.0\% & 0.0\% & 0.0\% & 0.0\% & 3.0\% & 2.6\% & 3.3\% & 19.5\% & 3.3\% & 18.6\% & 3.3\% & 18.8\% \\
        \hline

        SCAn & 83.0\% & 6.8\% & 96.0\% & 5.3\% & 100.0\% & 4.2\% & 96.7\% & 5.9\% & 96.7\% & 4.8\% & 100.0\% & 5.2\% \\
        \hline
        PiDAn-2 & 92.0\% & 13.2\%  & 98.0\% & 12.6\%  & 100.0\% & 12.1\% & 96.7\% & 10.7\% & 100.0\% & 11.0\% & 100.0\% & 9.8\% \\
        \hline
        PiDAn-2.5 & 92.0\% & 8.9\%  & 98.0\% & 9.5\%  & 100.0\% & 8.4\% & 96.7\% & 7.9\% &96.7\% & 7.4\% & 100.0\% & 7.4\% \\
        \hline
        \textbf{PiDAn-3} & \textbf{88.0\%} & \textbf{2.1\%}  & \textbf{98.0\%} & \textbf{2.1\%}  & \textbf{100.0\%} & \textbf{1.6\%} & \textbf{96.7\%} & \textbf{5.2\%} & \textbf{96.7\%} & \textbf{5.5\%} & \textbf{100.0\%} & \textbf{4.0\%} \\
        \hline\hline
    \end{tabular}
    \label{detection rates}
    \vspace{-0.1in}
\end{table*}

\subsubsection{Poisoned samples identifying via K-means} 
\begin{figure}[t]
	\centering
    \subfloat[]{\includegraphics[scale=0.45]{{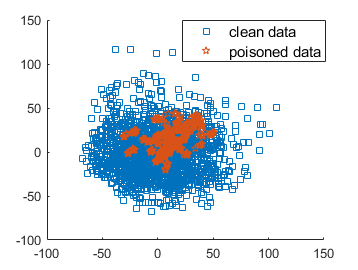}}\label{f:o1}}\
    \subfloat[]{\includegraphics[scale=0.45]{{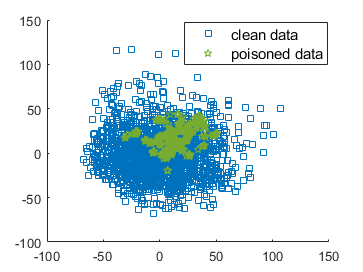}}\label{f:o2}}
%    \vspace{-0.2in}
    \caption{Top two features in the target class for the pair of speed limit of 30 and speed limit of 80 for hidden trigger backdoor attack. (a) ground truth; (b) identification results of our method. It is noted that our method can identify both the poisoned and authentic samples with high accuracy, e.g., more than 90\% poisoned samples.}
    \label{f:overlap}
    \vspace{-0.3in}
\end{figure}

It is important to identify the poisoned samples for further analyzing the backdoor attack and mitigating its influence in deep learning models after the infected classes are detected. Based on Section~\ref{ss:optimization}, the optimized weight vector for the infected class is further clustered into two classes by K-means and the clusters with less samples is considered to be the poisoned samples with high probability. We summarize the rates of located poisoned samples in Table~\ref{locate poisoned}. It is noted that our method can identify the poisoned with high accuracy and identify few genuine samples as the poisoned. For the case of similar source and target (speed limit 30 and speed limit 80), whose representations are shown in Fig~\ref{f:o1}, our method can achieve high poisoned sample identification accuracy, e.g., identify 90\% poisoned samples as shown in Fig~\ref{f:o2}, which is sufficient to suppress the backdoor influence after removing these poisoned samples.
%\vspace{-0.1in}
%
\begin{table*}[htpb]
    \centering
    \caption{The rates of identified poisoned and genuine samples for different detection methods on two datasets. Note that $\mathrm{TPR}$ denotes the rates of identifying poisoned samples and $\mathrm{FPR}$ is the false positive rates.}
    \vspace{-0.1in}
    %\small
    \begin{tabular}{c|c|c|c|c|c|c|c|c|c|c|c|c}
        \hline\hline
         \multirow{3}{*}{\tabincell{c}{Defense \\Method}} & \multicolumn{6}{c|}{ILSVARC2012} &\multicolumn{6}{c}{GTSRB}  \\
          \cline{2-13}
         &\multicolumn{2}{c|}{Hidden trigger}&\multicolumn{2}{c|}{TaCT}&\multicolumn{2}{c|}{Badnets} &\multicolumn{2}{c|}{Hidden trigger}&\multicolumn{2}{c|}{TaCT}&\multicolumn{2}{c}{Badnets}\\
         \cline{2-13}
            & TPR & FPR  & TPR & FPR & TPR & FPR & TPR & FPR  & TPR & FPR & TPR & FPR\\
         \hline
        AC \cite{chen2018detecting} & 74.6\% & 10.1\% & 74.9\% & 11.5\% & 80.3\% & 7.4\% & 82.4\% & 12.6\% & 82.9\% & 16.5\% & 71.4\% & 14.0\% \\
        \hline
        SCAn \cite{tang2021demon} & 64.8\% & 33.4\% & 65.1\% & 34.5\% & 65.8\% & 32.6\% & 33.1\% & 51.5\% & 33.4\% & 52.2\% & 35.0\% & 52.1\% \\
        \hline
        \textbf{PiDAn} & \textbf{95.8\%} & \textbf{4.4\%}  & \textbf{95.2\%} & \textbf{3.0\%} & \textbf{96.5\%} & \textbf{3.0\%} & \textbf{97.7\%} & \textbf{12.0\%} & \textbf{97.5\%} & \textbf{13.4\%} & \textbf{98.5\%} & \textbf{11.9\%} \\
        \hline\hline
    \end{tabular}
    \vspace{-0.1in}
    \label{locate poisoned}
\end{table*}
%
\begin{comment}
\begin{table*}[htpb]
    \centering
    \begin{tabular}{c|c|c|c|c|c|c|c|c}
        \hline\hline
         \multirow{3}{*}{Detection Method} & \multicolumn{4}{c|}{ILSVARC2012} &\multicolumn{4}{c}{GTSRB}  \\
         \cline{2-9}
            & \#clean target & \#poison  & \#poison identified & \#clean identified & \#clean target & \#poison  & \#poison identified & \#clean identified\\
         \hline
        Spectral Signature & 91.5\% & 65.2\% & 99.0\% & 80 \% &91.5\% & 65.2\% & 99.0\% & 80 \%\\
        \hline
        SCAn & 92.3\% & 85.2\% & 99.1\% & 80 &\%91.5\% & 65.2\% & 99.0\% & 80 \%  \\
        \hline
        Ours & 94.8\% & 86.7\%  & 99.5\% & 80\%  &91.5\% & 65.2\% & 99.0\% & 80 \% \\
        \hline\hline
    \end{tabular}
    \caption{The rates of located poisoned samples for different detection methods on two dataset.}
    \label{locate poisoned}
\end{table*}
\end{comment}
%
\subsubsection{Backdoor mitigation via poison samples exclusion}
As suggested in \cite{chen2018detecting}, after identifying the poisoned samples, we can remove these samples from the target class and retrain the model.
The performance of the repaired models are summarized in Table~\ref{mitigation}. It shows that the repaired models hold comparable classification accuracy since the test errors for the genuine samples are similar before and after removing the identified suspicious samples and repairing the backdoored models. Besides, the backdoor influences are successfully mitigated since less than 10\% of the source images carrying triggers are classified as the target. Specifically, for the case shown in Fig~\ref{f:overlap}, the test errors of the repaired model for the genuine and poisoned samples are 4.0\% and 9.5\% respectively.
\vspace{-0.1in}
\begin{table*}[htpb]
    \centering
    \caption{Test error for both the poisoned (P) and genuine (G) samples for different source-target pairs from ILSVRC2012 and GTSRB datasets before and after repairing the models.}
    \vspace{-0.1in}
    %\small
    \begin{tabular}{c|c|c|c|c|c|c|c|c|c|c|c|c|c|c}
        \hline \hline
        \multicolumn{2}{c|}{} & \multicolumn{10}{c|}{ILSVARC2012} & \multicolumn{3}{c}{GTSRB}\\
        \hline
        \multicolumn{2}{c|}{Pair ID} & 1 & 2 & 3 & 4 & 5 & 6 & 7 & 8 & 9 & 10 & 1 & 2 & 3\\
        \hline
        \multirow{2}{*}{Before} & G & 7.3\% & 7.5\% & 7.3\% & 7.3\% & 7.2\% & 7.3\% & 7.4\% & 7.6\% & 7.6\% & 7.3\% & 3.7\% & 4.0\% & 3.7\% \\
        \cline{2-15}
         & P & 93.1\% & 70.8\% & 82.3\% & 78.5\% & 89.2\% & 73.3\% & 86.8\% & 73.9\% & 91.1\% & 86.2\% & 97.4\% & 70.7\% & 97.1\%\\
        \hline
        \multirow{2}{*}{After} & G & 7.4\% & 7.5\% & 7.4\% & 7.4\% & 7.7\% & 7.6\% & 7.4\% & 7.7\% & 7.9\% & 7.5\% & 3.8\% & 4.0\% & 4.1\% \\
        \cline{2-15}
         & P & 7.7\% & 0.0\% & 2.3\% & 0.83\% & 12.3\% & 2.5\% & 8.8\% & 4.6\% & 3.7\% & 6.1\% & 5.1\% & 13.2\% & 7.9\% \\
        \hline \hline
    \end{tabular}
    \vspace{-0.2in}
    \label{mitigation}
\end{table*}

\subsection{Computation time}
\label{computation complexity}
We record the running time of the proposed PiDAn algorithm on ILSVRC2012 dataset and GTSRB dataset. For ILSVARC2012, it contains 20 classes and 16000 genuine samples and 400 poisoned samples, both with 4096 features. The running time of calculating the optimal weights is 124 seconds. For GTSRB, it contains 43 classes and 35288 genuine samples and 800 poisoned samples with 512 features. The running time is 6 seconds.
\vspace{-0.1in}
\section{Discussion}
%For each dataset, we report the average running time over different classes, which are 
%\vspace{-0.1in}
%
\subsection{Extension to multi-source cases} \label{multi-source}
\begin{figure}[htpb]
	\centering
     \subfloat[]{\includegraphics[scale=0.3]{{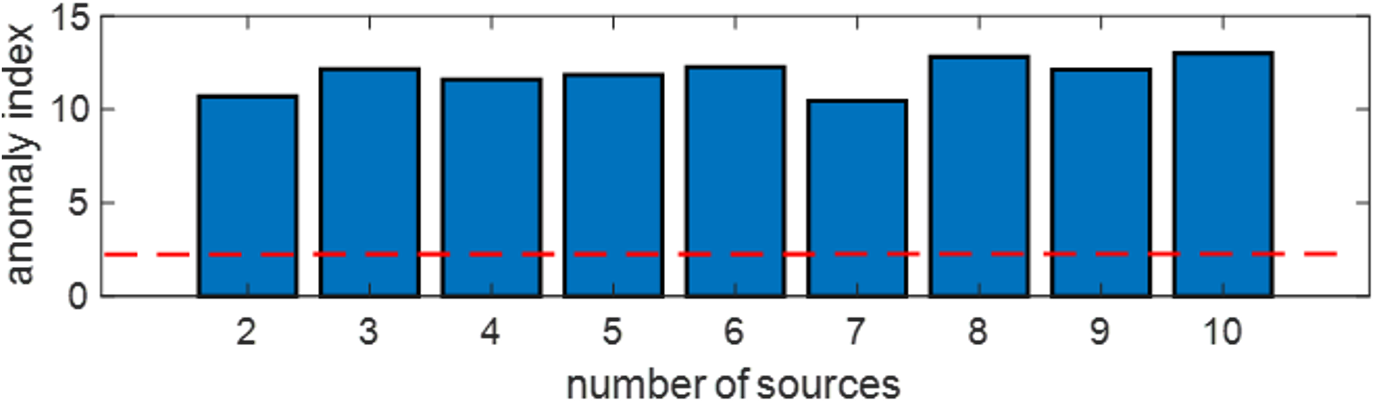}}}\ 
    \subfloat[]{\includegraphics[scale=0.3]{{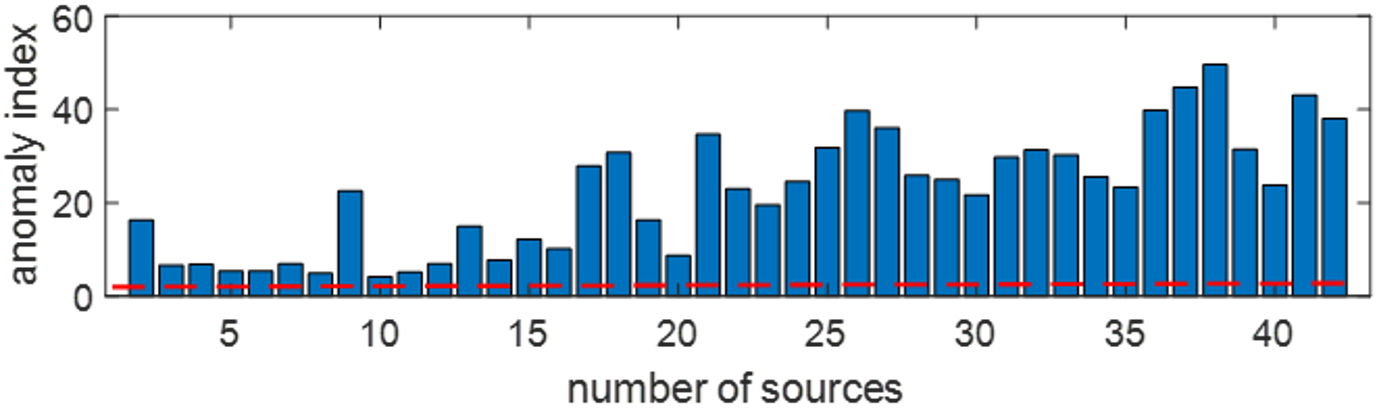}}}
    %\vspace{-0.1in}
    \caption{Anomaly index for (a) the ILSVARC2012 and (b) GTSRB datasets with different number of sources. The red dash line is the threshold 3 with 0.999 confidence level. }
    \label{f:multi_source}
    \vspace{-0.3in}
\end{figure}

We launch experiments on the ILSVRC2012 and GTSRB datasets to evaluate the effectiveness of our method on multi-source cases. For ILSVRC2012, we set the class \#20 as the target and vary the number of sources from 2 to 10; for GTSRB, the target class is set as the speed limit of 60 and the number of sources vary from 2 to 42. As shown in Fig~\ref{f:multi_source}, the anomaly indexes for all source numbers exceed the threshold 3, which indicates that our algorithm can detect the infected class with multiple sources. It is natural since our method aims to detect the infected classes by determining if they contain only genuine data. 
\vspace{-0.1in}

\begin{comment}
\begin{table}[htpb]
    \centering
    \caption{Anomaly index on ILSVARC2012 dataset for different numbers of sources. }
    \vspace{-0.1in}
    \begin{tabular}{c|c|c|c|c}
        \hline \hline
        sources ID & 1,2 & 1,2,3 & 1,2,3,4 & 1,2,3,4,5  \\
        \hline
        \# of poisoned & 200 & 300 & 400 & 500  \\
        \hline
        Anomaly index & 3.92 & 3.09 & 4.33 & 6.54 \\
        \hline \hline
    \end{tabular}
    \label{t:vs}
    \vspace{-0.2in}
\end{table}
\end{comment}

\begin{comment}
\begin{figure}[h!]
	\centering
     \includegraphics[scale=0.6]{{Imgs/vs.png}}
    \caption{Anomaly index on ILSVARC2012 dataset for different numbers of sources. The red dot line is the threshold with 0.95 confidence level.}
    \label{f:vs}
\end{figure}
\end{comment}
\begin{comment}
\begin{table}[htpb]
    \centering
    \begin{tabular}{c|c|c|c|c|c}
        \hline \hline
        \# of sources & 2 & 3 & 4 & 5 & 10 \\
        \hline
        TPR & 90.0\% & 90.5\% & 99.0\% & 90.0\% & 90.0\% \\
        \hline
        FPR & 91.5\% & 7.9\% & 90.0\% & 90\% & 90\% \\
        \hline \hline
    \end{tabular}
    \caption{The backdoor detection performance on ILSVARC2012 dataset for different numbers of sources. ($\mathrm{TPR}$ is the rate of detecting the infected class and $\mathrm{FPR}$ is the rate of detecting clean class as the infected.)}
    \label{varying sources}
\end{table}
\end{comment}
%
\subsection{Robustness to number of clean testing samples}
Since the clean testing samples are in need for the proposed PiDAn algorithm to centralize the mixture data representations,  we try on different number of clean testing samples for centralization to check how the detection performance changes. For each trial, we randomly pick up certain number of clean testing samples and report the average performance over 10 times. To highlight the performance of our proposed algorithm under different clean testing samples, the threshold is fixed to be 3 (with confidence level $99.9\%$) and the clean sample rate (CSR), e.g., for the target class,  $\mathrm{CSR}=\frac{\textrm{No. of clean testing samples}}{\textrm{No. of total genuine training samples}}$, varies from $0.2\%$ to $10\%$. The results regarding detection rate (DR) are summarized in Table \ref{Noclea}, where we  can observe that the performance of our method does not change significantly when clean sample rate is no less than $0.6\%$ for GTSRB dataset and larger than $1\%$ for ILSVARC2012 dataset.
This demonstrates that our method is robust to the size of clean samples when it stays in a feasible range, which further implies our PiDAn algorithm can work well based on very limited information of clean data.
\vspace{-0.1in}
\begin{table*}[htpb]
    \centering
    \caption{ TPR on both ILSVARC2012 and GTSRB datasets with different clean sample rates (CSR) for three attack methods. TPR denotes the rates of detecting the infected class}
    \vspace{-0.1in}
    %\small
    \begin{tabular}{c|c|c|c|c|c|c|c|c|c|c}
        \hline \hline
        \multicolumn{11}{c}{ILSVARC2012} \\
        \hline
        CSR & 0.2\% & 0.4\% & 0.6\% & 0.8\% & 1\% & 2\% & 4\% & 6\% & 8\% & 10\%\\
        \hline\hline
        Hidden trigger & 61.0\% & 63.0\% & 68.0\% & 73.0\% & 81.0\% & 83.0\% & 85.0\% & 86.0\% & 88.0\% & 88.0\% \\
        \hline
        TaCT & 65.0\% & 70.0\% & 73.0\% & 78.0\% & 82.0\% & 86.0\% & 90.0\% & 92.0\% & 97.0\% & 98.0\%\\
        \hline
        Badnets &  63.0\% & 71.0\% & 75.0\% & 78.0\% & 81.0\% & 88.0\% & 93.0\% & 96.0\% & 98.0\% & 100\%\\
        \hline\hline
        \multicolumn{11}{c}{GTSRB} \\
        \hline
       CSR & 0.2\% & 0.4\% & 0.6\% & 0.8\% & 1\% & 2\% & 4\% & 6\% & 8\% & 10\%\\
        \hline\hline
        Hidden trigger & 76.7\% & 83.3\% & 90.0\% & 93.3\% & 93.3\% & 96.7\% & 96.7\% & 100\% & 96.7\% & 96.7\% \\
        \hline
        TaCT & 80.0\% & 86.7\% & 90.0\% & 90.0\% & 93.3\% & 93.3\% & 96.7\% & 100\% & 96.7\% & 100\% \\
        \hline
        Badnets & 86.7\% & 90.0\% & 93.3\% & 96.7\% & 96.7\% & 100\% & 96.7\% & 96.7\% & 100\% & 96.7\% \\
        \hline \hline
    \end{tabular}
    \vspace{-0.3in}
    \label{Noclea}
\end{table*}
\subsection{Performance on different poisoned sample rates} 
\begin{figure}[htpb]
	\centering
    \subfloat[]{\includegraphics[scale=0.45]{{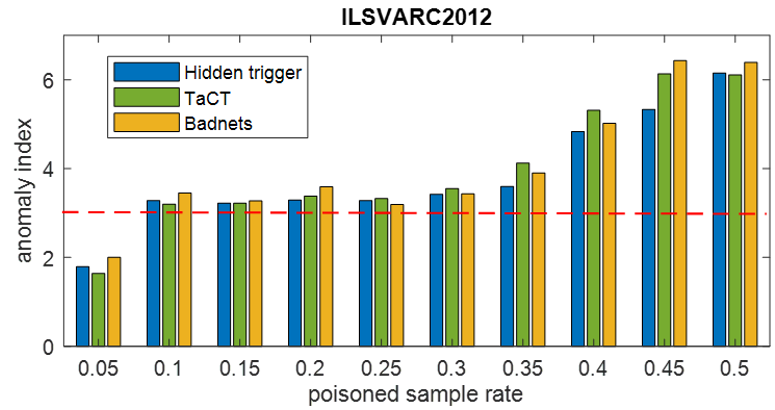}}}
    \hfil
%    \vspace{-0.1in}
    \subfloat[]{\includegraphics[scale=0.45]{{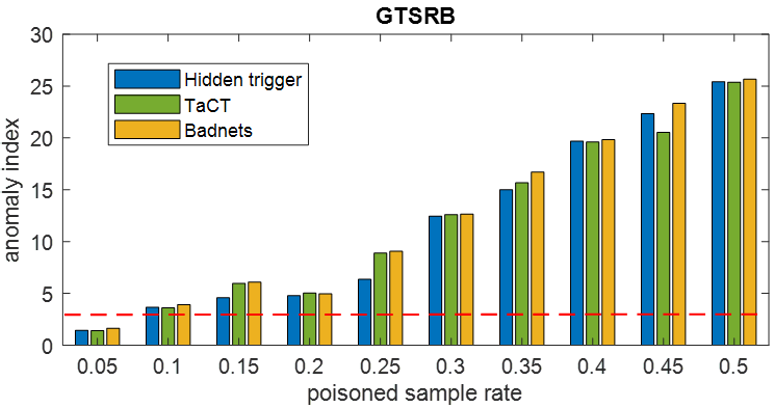}}}
%    \vspace{-0.1in}
    \caption{Anomaly index for different infected models for (a) the ILSVARC2012 and (b) GTSRB dataset under different attacks (blue, green and yellow respectively) with varying poisoned sample rate. The red dash line is the threshold 3 with 0.999 confidence level. (Poisoned sample rate $= \frac{\textrm{No. of poisoned samples}}{\textrm{No. of genuine training samples}}$)}
    \label{f:vpn}
    \vspace{-0.2in}
\end{figure}

%Here, we demonstrate that our PiDAn algorithm is robust to the number of  poisoned samples, \emph{a.k.a.}, poisoned sample rate. To illustrate this, we try different number of injected poisoned samples on the three infected models for the ILSVARC2012 and GTSRB source-target pairs under different attack strategies, e.g., Hidden trigger attack, TaCT, and Badnets. For all attack strategies, the attack success rates increase with more poisoned samples while the classification rates keep in a high level of above 90\%. We present the anomaly index of different source-target pairs under different poisoned sample rate in  Fig~\ref{f:vpn}, from which it is noted that, under different attack strategies, all the anomaly indexes stay above the thresholds with different settings, which implies our proposed PiDAn algorithm can successfully detect infected class. This demonstrates our proposed PiDAn's robustness to the poisoned sample rate.
Here, we evaluate our PiDAn algorithm for different numbers of  poisoned samples, \emph{a.k.a.}, poisoned sample rate, where poisoned sample rate $= \frac{\textrm{No. of poisoned samples}}{\textrm{No. of genuine training samples}}$. To illustrate this, we try different number of injected poisoned samples for the ILSVARC2012 and GTSRB dataset under different attack strategies, e.g., Hidden trigger attack, TaCT, and Badnets. For all the attack strategies, the attack success rates increase with the number of poisoned samples while the classification rates keep in a high level of above 90\%. We present the anomaly index under different poisoned sample rate in Fig~\ref{f:vpn}. We observe that, when the poisoned sample rate exceeds 10\%, all the anomaly indices stay above the detection threshold with different settings under different attack strategies. This implies our proposed PiDAn algorithm can successfully detect infected class. From the figure we also observe that for the 5\% poisoned sample rate, the anomaly index is $\approx 2$ and thus, is below our threshold 3. Lower poisoning rates (for example: 5\% in case of ILSVARC and GTSRB datasets) may evade detection schemes as also reported by other researchers \cite{tran2018spectral}, and an attacker may choose a lower poisoning rate. However, the effectiveness of the attack significantly reduces with the decrease in poisoning rate as we see that the attack success rates for backdoor attacks drop to 65\% for ILSVARC2012 and 80\% for GTSRB. Backdoor attacks are considered dangerous since they are typically characterized with very high attack success rates (usually $>$95\%)\cite{gu2017badnets}. To keep the attack success rate high, advanced/stealthier attacks have proposed poisoning rate as high as 50\% \cite{liu2019abs, sarkar2020facehack}.
\vspace{-0.2in}
%For backdoor detection, the infected detection rates for our method change little with the change of the injected poisoned samples. Besides, our method still hold the high accuracy of identifying poisoned samples with only 5\% clean samples mis-identified as the poisoned.
%\vspace{-0.05in}
%
\subsection{Performance on varying trigger properties} \label{varying size and locations}
\begin{figure}[t]
    \centering
    \subfloat[]{\includegraphics[scale=0.2]{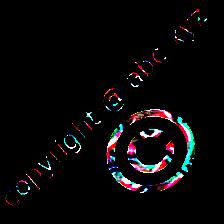}}
    \hfil
    %
%    \vspace{-0.1in}
    \subfloat[]{\includegraphics[scale=0.36]{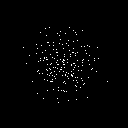}}
    %
%    \vspace{-0.1in}
    \caption{Two triggers used in our experiments.}
    \label{f:more_trigger}
    \vspace{-0.1in}
\end{figure}

\begin{figure}[t]
    \centering
    \includegraphics[width=0.4\textwidth]{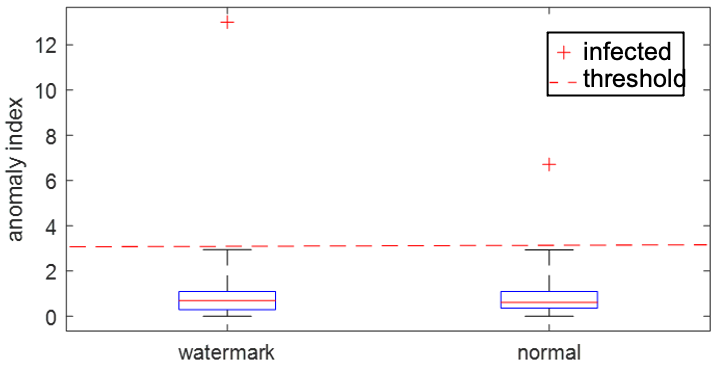}
    \vspace{-0.1in}
    \caption{Anomaly index for watermark and normal trigger. Red dash line is the threshold 3 with 0.999 confidence level.}
    \vspace{-0.1in}
    \label{f:more_boxplot}
\end{figure}

For the GTSRB dataset, we vary the trigger size (from 3*3 to 7*7) and location (four corners and the middle). We calculate the anomaly index for each size or location and summarize the results in the Table \ref{t:vtp}. In Section \ref{exp}, we evaluate PiDAn on colored square triggers. We also evaluate it on different trigger shapes for the Badnets scheme, e.g., watermark and normal trigger in Fig.\ref{f:more_trigger}, which cover the whole image. As shown in Fig~\ref{f:more_boxplot}, the anomaly index of infected classes for the watermark and normal trigger all exceed 3, which shows that our algorithm can detect the backdoor in the model as well as the infected class. 
\vspace{-0.1in}
\begin{table}[htpb]
    \centering
    \caption{Anomaly index of the infected class for different trigger sizes and positions.}
    \vspace{-0.1in}
    %\small
    \begin{tabular}{c|c|c|c|c|c}
        \hline \hline
        sizes & 3*3 & 4*4 & 5*5 & 6*6 & 7*7  \\
        \hline
        \tabincell{c}{Anomaly \\index} & 12.96 & 14.58 & 10.04 & 11.46 & 12.67 \\
        \hline \hline
       Positions & \tabincell{c}{left\\top} & \tabincell{c}{right\\top} & \tabincell{c}{left\\bottom} & \tabincell{c}{right\\bottom} & middle \\
        \hline
        \tabincell{c}{Anomaly \\index} & 17.94 & 14.06 & 14.26 & 10.43 & 13.08 \\
        \hline \hline
    \end{tabular}
    \label{t:vtp}
    \vspace{-0.2in}
\end{table}
\vspace{-0.1in}
\subsection{Performance on Binary classifiers} 
We also perform the proposed PiDAn algorithm on Binary classifiers, which are trained by our source-target pairs from the ILSVARC-2012 and GTSRB dataset and the target classes are infected by poisoned samples. We summarize the likelihood ratio for both the clean source classes and the infected target classes in Table \ref{binary}. The larger the likelihood ratio, the higher probability that the corresponding class is infected. We can see that the likelihood ratio for the infected target class is always much larger than that of the clean source class, indicating that our proposed PiDAn algorithm has the ability of distinguishing the clean and the infected classes.
\vspace{-0.1in}
\begin{table*}[htpb]
    \centering
    \caption{Likelihood ratios for different source (S) and target (T) pairs from ILSVRC2012 and GTSRB datasets.}
    \vspace{-0.1in}
    %\small
    \begin{tabular}{c|c|c|c|c|c|c|c|c|c|c|c|c|c}
        \hline \hline
        \multicolumn{1}{c|}{} & \multicolumn{10}{c|}{ILSVARC2012} & \multicolumn{3}{c}{GTSRB}\\
        \hline
        Pair ID & 1 & 2 & 3 & 4 & 5 & 6 & 7 & 8 & 9 & 10 & 1 & 2 & 3\\
        \hline
        S & 37.86 & 26.33 & 45.73 & 104.74 & 67.15 & 33.31 & 117.16 & 93.47 & 55.58 & 90.94 & 56.89 & 118.85 & 117.14 \\
        \cline{1-14}
        T & 474.60 & 514.99 & 423.38 & 480.78 & 359.76 & 701.42 & 818.21 & 734.94 & 491.59 & 633.65 & 970.46 & 629.99 & 1103.8\\
        \hline \hline
    \end{tabular}
    \vspace{-0.1in}
    \label{binary}
\end{table*}
\begin{table*}[htpb]
    \centering
    \caption{Number of removed poisoned (RP) samples and test error for both the poisoned (P) and genuine (G) samples for different source-target pairs from ILSVRC2012 and GTSRB datasets after repairing the models by Spectral signature. (The total number of injected poisoned samples for ILSVARC2012 is 400 and that for GTSRB is 800).}
    \vspace{-0.1in}
    %\small
    \begin{tabular}{c|c|c|c|c|c|c|c|c|c|c|c|c|c}
        \hline \hline
        \multicolumn{1}{c|}{} & \multicolumn{10}{c|}{ILSVARC2012} & \multicolumn{3}{c}{GTSRB}\\
        \hline
        \multicolumn{1}{c|}{Pair ID} & 1 & 2 & 3 & 4 & 5 & 6 & 7 & 8 & 9 & 10 & 1 & 2 & 3\\
        \hline
         RP & 0 & 100 & 87 & 126 & 74 & 72 & 157 & 163 & 56 & 174 & 105 & 23 & 215 \\
        \cline{1-14}
         Test error-G & 7.5\% & 7.0\% & 7.3\% & 7.6\% & 7.5\% & 7.4\% & 7.4\% & 7.7\% & 7.8\% & 7.9\% & 3.8\% & 4.6\% & 4.4\% \\
        \cline{1-14}
         Test error-P & 96.6\% & 70.0\% & 73.9\% & 68.6\% & 86.9\% & 71.7\% & 67.7\% & 54.6\% & 93.1\% & 83.1\% & 93.6\% & 76.9\% & 75.7\%\\
        \hline
        \hline
    \end{tabular}
    \vspace{-0.2in}
    \label{spectral signature}
\end{table*}
\subsection{Performance on adaptive attacks} 
\begin{figure}
    \centering
    \includegraphics[scale=0.55]{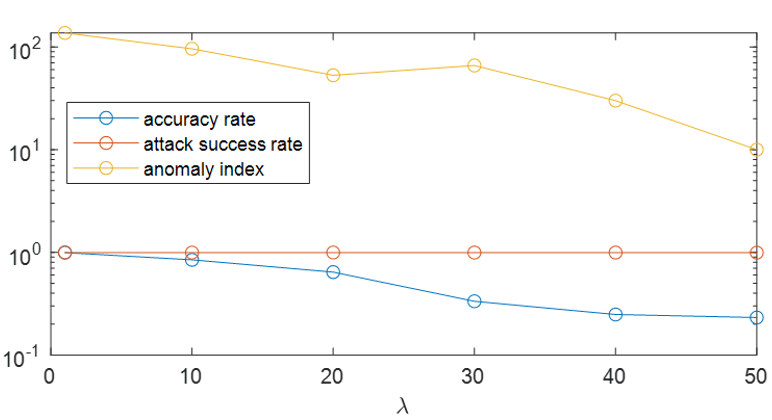}
    \vspace{-0.1in}
    \caption{The accuracy rate, attack success rate for backdoored models and anomaly index for infected classes  with varying weight ($\lambda$) for the loss of the discriminative network in adversarial embedding attack. The accuracy rate decreases with the increase of $\lambda$, and the anomaly index also decreases but is still larger than the threshold 3, showing that our method can detect the infected classes.}
    \vspace{-0.2in}
    \label{f:adaptive attack}
\end{figure}

As proposed in \cite{shokri2020bypassing}, it is important to evaluate the proposed defense algorithm on the adaptive attacks where the attacker knows the defenses and tries to bypass them. In~\cite{shokri2020bypassing}, an adversarial embedding attack is proposed, where the differences between the poisoned and genuine representations are minimized so that defenses based on representation distinguishing could be bypassed. To achieve this, they modify the original loss function by adding a classification accuracy based loss which reveals whether representations correspond to correct classes, i.e., poisoned or genuine.  We implement the adversarial embedding attack via the Adversarial Robustness toolbox \cite{nicolae2018adversarial}.
%, and perform the experiments using the same datasets, the Cifar10 and GTSRB, on the same model architectures, DenseNet and VGG. Firstly, we follow the same parameters in the paper and our algorithm can detect the poisoned classes in all cases and the anomaly indexes for these 4 experiments are summarized in Table \ref{t:adaptive attack}.
We use the same experiment setting in~\cite{shokri2020bypassing}, injecting a backdoor trigger in 5\% of training samples and setting their labels to their arbitrarily chosen target label $y_t = 2$. We increase the weight ($\lambda$) for the loss of the discriminative network, that is, force the genuine and poisoned representations to be closer, so that the ability of the adversarial embedding attack to bypass our algorithm can increase. Fig. \ref{f:adaptive attack} shows that with the increase of the weight, the accuracy of the backdoored model decreases heavily, e.g., around 20\% for $\lambda=40$ and $\lambda = 50$, while our algorithm can still detect the poisoned classes since the anomaly index of the poisoned class for each $\lambda$ exceeds the threshold 3.
\vspace{-0.1in}
\begin{comment}
\begin{table}[]
    \caption{Anomaly index for experiments from adversarial embedding attack}
    \centering
    \begin{tabular}{c|c|c}
    \hline \hline
         & Cifar10 & GTSRB  \\ \hline
    DenseNet &  &  \\ \hline
    VGG &  &  \\
    \hline \hline
    \end{tabular}
    \label{t:adaptive attack}
\end{table}
\end{comment}
%
\begin{comment}
\begin{table}[]
    \caption{Anomaly index for experiments from adversarial embedding attack}
    \centering
    \begin{tabular}{c|c|c|c|c|c|c}
    \hline \hline
     $\lambda$ & 1 & 10 & 20 & 30 & 40 & 50  \\ \hline
    accuracy rate & 0.993 & 0.845 & 0.642 & 0.334 & 0.248 & 0.232  \\ \hline
    attack success rate & 0.997 & 0.994 & 0.996 & 0.994 & 0.996 & 0.992\\ \hline
    attack detection rate & 137.05 & 96.00 & 53.08 & 66.49 & 30.67 & 9.56 \\ 
    \hline \hline
    \end{tabular}
    \label{t:adaptive attack}
\end{table}
\end{comment}
%
\subsection{The selection of \textit{k}}
In our algorithm, we first extract the latent subspace of the training data, which is the top $k$ components from principal component analysis (PCA). The selection of $k$ can resort to several methods, such as the cross validation and cumulative percentage variance~ \cite{li2000recursive,li2021nonlinear}. In this work, we determine the number of components in $\mathbf{P}$, i.e., the eigenvectors, by the cumulative percentage variance (CPV)~\cite{li2000recursive}, which is a measure of how much variation is captured by the first $k$ components and is
\begin{equation}
    \mathrm{CPV}(k)=\frac{\sum^k_{i=1}\lambda_i}{\sum_n^{i=1}\lambda_i}\times 100\%
    \label{cpv}
\end{equation}
where $\lambda_i$ is the $i$th eigenvalue describing the variation of the $i$th component in $\mathbf{P}$. The number of components, a.k.a., the parameter $k$, is selected when CPV reaches the threshold, say 95\%, where the statistical variation of data is captured with noise excluded.

We also experiment with different thresholds of CPV from 95\% to 99\% on the GTSRB dataset and summarize the anomaly index for infected class in Table \ref{f:CPV}. We can see that the anomaly index for different CPV are almost the same, indicating that our backdoor detection algorithm does not rely on the threshold of CPV when it exceeds 95\%, since major variation is captured. 
\vspace{-0.1in}
\begin{table}[htpb]
    \centering
    \caption{Backdoor detection rate on GTSRB datasets with cumulative percentage variation (CPV).}
    \vspace{-0.1in}
    %\small
    \begin{tabular}{c|c|c|c|c|c}
        \hline \hline
        CPV & 95\% & 96\% & 97\% & 98\% & 99\% \\
        \hline
        Anomaly index & 35.45 & 37.52 & 36.89 & 34.51 & 35.22 \\
        \hline \hline
    \end{tabular}
    \vspace{-0.2in}
    \label{f:CPV}
\end{table}
\section{Comparison with state-of-the-art}
In this section, we compare our method with three state-of-the-art methods which, similar to our method, defend against backdoor attacks in neural networks by distinguishing poisoned samples from the clean ones.
\vspace{-0.1in}
\subsection{Spectral signatures}
Spectral signatures \cite{tran2018spectral} identifies the poisoned samples by the top principals of corrupted data. Their intuition is that if the means of the authentic and poisoned representations are sufficiently well-separated relative to their variance, the correlations of the poisoned samples with the top principal component of the corrupted data via singular value decomposition are large in magnitude. Then the backdoor can be removed by excluding a fixed number of samples with large correlations. We remove 15\% samples with top correlations as suggested in~\cite{tran2018spectral}. In this manner, Spectral signatures can remove less than 50\% of the poisoned samples for ILSVARC2012 and less than 30\% for GTSRB. After removing the suspicious samples and retraining the backdoored models, the repaired models still hold high Attack Success rates as shown in Table \ref{spectral signature}, which is not sufficient to mitigate the influence of the backdoor.
\vspace{-0.1in}
\subsection{Activation clustering}
AC~\cite{chen2018detecting} uses a similar strategy to Spectral signatures. It is based on the idea that the representations of the data in the infected class can be separated into two clusters, the poisoned and the authentic ones, by clustering the top principal components by K-means. They cluster each class to two groups and the one with significantly large Silhouette Score is considered to be infected since two clusters better describe this class. In their paper \cite{chen2018detecting}, the authors recommend to set a threshold between 0.10 and 0.15 and the class with the Silhouette Score larger than this threshold will be detected as the infected class. We use AC on the infected representations in our experiments and report the detection performance in Table \ref{detection rates} with different thresholds (0.10, 0.11, 0.12, 0.13, 0.14, 0.15). AC can detect more than 50\% infected classes when the threshold is 0.10, but it also introduces large false positive, e.g., wrongly detecting around 50\% clean classes as the infected; while for larger thresholds, AC loses efficiency of detecting the infected classes. The reason AC fails lies on that the Silhouette Score of the clean and infected classes are in the same scale e.g., both ranging from 0.08 to 0.14, so that it has issues detecting the infected class by setting a threshold and determining those beyond the threshold as infected. Furthermore, we investigate its capability of identifying the genuine and poisoned samples in the infected classes in Table~\ref{locate poisoned}, assuming that it can first successfully detect them. We compute true positive rate (TPR) and false positive rate (FPR), in order to evaluate the performance of identifying genuine samples and poisoned samples, respectively. As summarized in Table \ref{locate poisoned}, our method has larger TPR (e.g., larger than $95.2\%$ ) and FPR (e.g., as small as $3.0\%$) than AC (e.g., with TPR no more than $82.9\%$ and FPR around $10\%$), which implies our method exceeds the  performance of AC.
\vspace{-0.1in}
\subsection{SCAn}
SCAn provides a strategy to distinguish the distributions of the poisoned and the authentic. It assumes that the representations of poisoned data and authentic data in the infected classes both follow Gaussian distributions with the same variance and different means. They propose an iterative algorithm to untangle the data to two groups for each class and establish a test statistic based on the likelihood ratio. The test statistic for some class exceeding the threshold indicates the high probability that the data in this class follow two Gaussian distributions mixture, which indicates this class is infected. We also evaluate this method on our infected representations. As shown in Table~\ref{detection rates}, SCAn achieves  comparable infected class detect rates as our method, however, with many more clean classes identified as the infected. As shown in Table~\ref{locate poisoned}, SCAn can only identify around 60\% poisoned samples with around 40\% authentic ones identified as the poisoned. It cannot correctly cluster the authentic and poisoned data into two subgroups in our cases and cannot help repair the infected models. Besides, the running time for the dataset mentioned in Section \ref{computation complexity} are 11.2 hours for ILSVARC2012 dataset and 550 seconds for GTSRB dataset respectively, which are much longer than our PiDAn algorithm.
%\vspace{-0.1in}

\section{Conclusion}
This paper has presented a novel backdoor attack detection and mitigation method by solving a coherence optimization problem. The efficacy of our method has been theoretically and empirically illustrated. On the theoretical side, we prove that the weight vector (obtained by coherence optimization formulation) has a ``grouping effect'' and is different for authentic data and poisoned data, which enables the detection and mitigation of backdoor attacks. At the same time, extensive experiments against three state-of-the-art backdoor schemes, including the emerging hidden trigger backdoor attack, validate the effectiveness and robustness of our method in detecting the infected classes. Our method can successfully identify more than 95\% poisoned samples with less than 10\% authentic samples identified as poisoned. Therefore, our method can also help mitigate the backdoor influence by correctly removing the poisoned samples from the training dataset and also maintain high classification accuracy. 
\label{con}

%\section*{Acknowledgment}
%This work was jointly supported by the NYUAD Center for Interacting Urban Networks (CITIES), funded by Tamkeen under the NYUAD Research Institute Award CG001 and by the Swiss Re Institute under the Quantum Cities™ initiative, and Center for CyberSecurity (CCS), funded by Tamkeen under the NYUAD Research Institute Award G1104.

\section*{Open Source}

All the source code and dataset proposed will be open-sourced after the review process.

\begin{comment}
\section*{ACKNOWLEDGMENT}
This work was jointly supported by the NYUAD Center for Interacting Urban Networks (CITIES), funded by Tamkeen under the NYUAD Research
Institute Award CG001 and by the Swiss Re Institute under the Quantum Cities™ initiative, and Center for CyberSecurity (CCS), funded by Tamkeen under the NYUAD Research Institute Award G1104.
\end{comment}

\bibliographystyle{IEEEtran}
\bibliography{references.bib}
%\vspace{12pt}

\begin{comment}
\section{Supplementary Materials}
\begin{figure*}[h!]
	\centering
    \subfigure[]{ \includegraphics[scale=0.45]{{Imgs/m1.png}}}
    \subfigure[]{ \includegraphics[scale=0.45]{{Imgs/m2.png}}}
    \subfigure[]{ \includegraphics[scale=0.45]{{Imgs/m3.png}}}
    \subfigure[]{ \includegraphics[scale=0.45]{{Imgs/m4.png}}}
        \subfigure[]{ \includegraphics[scale=0.45]{{Imgs/m5.png}}}

    \caption{Examples of overlapped neuron representations in  ILSVARC2012 dataset. The representations are projected onto the space spanned by first two principal directions. The top figures denote the ground truth; The middle ones are our identification results; The bottom ones are optimized coefficients.}
    \label{f:mm}
\end{figure*}
\end{comment}

\appendices

\section{Proof of Remark 3}
\label{Ap2}
For the optimization problem in Eq.\eqref{Op-1}, the Lagrangian function is
\begin{equation}
    \mathcal{L}(\mathbf{a})=\mathbf{a}^\top(\mathbf{\Omega}_2-\mathbf{\Omega}_1)\mathbf{a}-\lambda\mathbf{a}^\top\mathbf{a}
\end{equation}

Since $\mathbf{a}^*$ is the optimal solution of Eq.\eqref{Op-1}, we have
\begin{equation}
\frac{\partial\mathcal{L}(\mathbf{a})}{\partial \mathbf{a}}|_{\mathbf{a}=\mathbf{a}^*}=0
\end{equation}

This gives
\begin{equation}
\begin{split}
&\mathbf{y}^\top_i\mathbf{Y}\mathbf{a}^*-\lambda^*\mathit{a}^*_i=0\\
&\mathbf{y}^\top_j\mathbf{Y}\mathbf{a}^*-\lambda^*\mathit{a}^*_j=0
\end{split}
\label{eq-pro2}
\end{equation}
where, $\mathbf{y}_i$ is the $i$th row of $\mathbf{Y}$ and $\mathit{a}^*_i$ is the $i$th element in $\mathbf{a}^*$. From Eq.\eqref{eq-pro2}, we further have,
\begin{equation}
    \mathit{a}^*_i-\mathit{a}^*_j=\frac{1}{\lambda^*}(\mathbf{y}^\top_i-\mathbf{y}^\top_j)\mathbf{Y}\mathbf{a}^*
\end{equation}

Since $||\mathbf{Y}\mathbf{a}^*||=\sqrt{\lambda^*}$ and $||\mathbf{y}^\top_i-\mathbf{y}^\top_j||=||(\mathbf{x}^\top_i-\mathbf{x}^\top_j)(\mathbf{I}-\mathbf{P}\mathbf{P}^\top)||\leq||\mathbf{x}^\top_i-\mathbf{x}^\top_j||\cdot||\mathbf{I}-\mathbf{P}\mathbf{P}^\top||\leq||\mathbf{x}^\top_i-\mathbf{x}^\top_j||$, we have
\begin{equation}
    \begin{split}
        |\mathit{a}^*_i-\mathit{a}^*_j|\leq\frac{1}{\sqrt{\lambda^*}}||\mathbf{y}^\top_i-\mathbf{y}^\top_j||
        &\leq\frac{1}{\sqrt{\lambda^*}}||\mathbf{x}^\top_i-\mathbf{x}^\top_j||\\
        &\leq\sqrt{\frac{2(1-\rho_{ij})}{\lambda^*}}
    \end{split}
\end{equation}
when data points are normalized. This completes the proof.
$\hfill\square$

%\section{Model architecture for GTSRB}
%The architecture of the 6 convolution layers and 2 dense layers neural network for the GTSRB dataset is summarized in Table~\ref{architecture for GTSRB}.
\begin{table}[ht]
    \centering
    \caption{Model architecture for GTSRB}
    \vspace{-0.1in}
    \footnotesize
    \begin{tabular}{c c c c c}
    \hline
    \hline
     Layer Type & \# of Channels & Filter Size & Stride & Activation \\
    \cline{1-5}
      Conv  & 32 & 3$\times$3 & 1 & ReLU \\
      Conv  & 32 & 3$\times$3 & 1 & ReLU \\
      MaxPool  & 32 & 2$\times$2 & 2 & -- \\
      Conv  & 64 & 3$\times$3 & 1 & ReLU \\
      Conv  & 64 & 3$\times$3 & 1 & ReLU \\
      MaxPool  & 64 & 2$\times$2 & 2 & -- \\
      Conv  & 128 & 3$\times$3 & 1 & ReLU \\
      Conv  & 128 & 3$\times$3 & 1 & ReLU \\
      MaxPool  & 128 & 2$\times$2 & 2 & -- \\
      FC  & 512 & -- & -- & ReLU \\
      FC  & 43 & -- & -- & Softmax \\
    \hline
    \hline
    \end{tabular}
    \vspace{-0.1in}
    \label{architecture for GTSRB}
\end{table}
\section{Definition of flattening metric}
\label{ap4}
The flattening metric is derived based on the similarity between Euclidean distance and Geodesic distance that is approximated by graph distance. By forming up a nearest neighborhood graph for all samples, the Geodesic distance between two points can be approximated by the sum of the length of edges included in the shortest path.

Let $G(i,j)$ denotes the Geodesic distance between sample $\mathbf{x}_i$ and $\mathbf{x}_j$, we have
\begin{equation}
    G(i,j)=\sum_{k}|e_k|
\end{equation}
where $|e_k|$ is the length of $k$th edge included in the shortest path.
The Euclidean distance between sample $\mathbf{x}_i$ and $\mathbf{x}_j$ is given by
\begin{equation}
    E(i,j)=||\mathbf{x}_i-\mathbf{x}_j||_2
\end{equation}
Let $\mathbf{r}_G$ and $\mathbf{r}_E$ denote the vectors containing Geodesic distance and Euclidean distance for all pairs of points, respectively. The flattening metric is defined as
\begin{equation}
   c=1-\frac{2}{N(N-1)}\mathbf{\tilde{r}}_G^\top\mathbf{\tilde{r}}_E
\end{equation}
where $\mathbf{\tilde{r}}_G$ and $\mathbf{\tilde{r}}_E$ denote the normalized $\mathbf{r}_G$ and $\mathbf{r}_E$, and $N$ is the number of all samples.

\end{document}